\documentclass{ieeetj}
\usepackage{cite}
\usepackage{amsmath,amssymb,amsfonts}
\usepackage{algorithmic}
\usepackage{graphicx,color}
\usepackage{textcomp}
\usepackage{xcolor}
\usepackage{hyperref}
\hypersetup{hidelinks=true}
\usepackage{algorithm,algorithmic}
\usepackage{etoolbox}

\def\BibTeX{{\rm B\kern-.05em{\sc i\kern-.025em b}\kern-.08em
    T\kern-.1667em\lower.7ex\hbox{E}\kern-.125emX}}
\AtBeginDocument{\definecolor{tmlcncolor}{cmyk}{0.93,0.59,0.15,0.02}\definecolor{NavyBlue}{RGB}{0,86,125}}

\def\authorrefmark#1{\ensuremath{^{\textbf{#1}}}}

\def\OJlogo{}
\def\seclogo{}

\makeatletter
\renewcommand{\receiveddate}[1]{}
\renewcommand{\reviseddate}[1]{}
\renewcommand{\accepteddate}[1]{}
\renewcommand{\publisheddate}[1]{}
\renewcommand{\currentdate}[1]{}
\renewcommand{\doiinfo}[1]{}
\makeatother

\begin{document}


\title{A B-Spline Function based 3D Point Cloud Unwrapping Scheme for 3D Fingerprint Recognition and Identification}

\author{M. Mogharen Askarin \authorrefmark{1}, J. Hu\authorrefmark{1}, M. Wang\authorrefmark{2},X. Yin\authorrefmark{3},\\ and X. Jia.\authorrefmark{4}}
\affil{School of Systems and Computing, UNSW Canberra, ACT 2600, Australia}
\affil{Information Systems, University of Canberra, ACT 2617, Australia}
\affil{School of Information and Communication Technology, Griffith University, Queensland 4222, Australia}
\affil{School of Engineering \& Technology, UNSW Canberra, ACT 2600, Australia}
\corresp{Corresponding author: Prof. J. Hu (email: J.Hu@adfa.edu.au).}

\begin{abstract}
Three-dimensional (3D) fingerprint recognition and identification offer several advantages over traditional two-dimensional (2D) recognition systems. The contactless nature of 3D fingerprints enhances hygiene and security, reducing the risk of contamination and spoofing. In addition to surface ridge and valley patterns, 3D fingerprints capture depth, curvature, and shape information, enabling the development of more precise and robust authentication systems. Despite recent advancements, significant challenges remain. The topological height of fingerprint pixels complicates the extraction of ridge and valley patterns. Furthermore, registration issues limit the acquisition process, requiring consistent direction and orientation across all samples. To address these challenges, this paper introduces a method that unwraps 3D fingerprints, represented as 3D point clouds, using B-spline curve fitting to mitigate height variation and reduce registration limitations. The unwrapped point cloud is then converted into a grayscale image by mapping the relative heights of the points. This grayscale image is subsequently used for recognition through conventional 2D fingerprint identification methods. The proposed approach demonstrated superior performance in 3D fingerprint recognition, achieving Equal Error Rates (EERs) of 0.2072\%, 0.26\%, and 0.22\% across three experiments, outperforming existing methods.  Additionally, the method surpassed 3D fingerprint flattening technique in both recognition and identification during cross-session experiments, achieving an EER of 1.50\% when fingerprints with varying registrations were included.

\end{abstract}

\maketitle

\section{INTRODUCTION} 
A 3D fingerprint is a three-dimensional representation of a fingerprint, capturing not only the ridge and valley patterns typically seen in traditional 2D fingerprint images but also the depth and curvature information of the finger's surface~\cite{kumar2018contactless}. Conventional fingerprint scanners produce 2D images, which can be vulnerable to spoofing and replication through the use of fingerprint molds~\cite{galbally2013image,cao2016hacking,askarin2018planting}. This extra information provides a more complete and accurate representation compared to a 2D fingerprint, making it useful for more reliable fingerprint recognition and identification which has applications in diverse areas, including biometric authentication, border control, and law enforcement~\cite{jain2004introduction}.

A 3D fingerprint can be represented as a three-dimensional point cloud. This point cloud consists of a collection of data points in 3D space that captures the detailed surface structure of a fingerprint. Each point in the cloud is defined by three coordinates ($x$,$y$,$z$), which specify its precise location and height on the fingerprint's surface.

A 3D fingerprint point cloud can be created using cameras and specialized scanning devices, such as laser scanners, structured light scanners, and photogrammetry systems. These devices reconstruct the depth information of the fingerprint surface by employing techniques like triangulation and projection~\cite{kumar2018contactless,labati2015toward,wang2010data,kumar2013towards,sousedik2013volumetric,tang20163}.
The raw data collected can be processed to create a point cloud representation of the fingerprint surface.

3D fingerprint technology addresses the problem of nonlinear elastic distortion commonly encountered in 2D contact-based fingerprint applications. This distortion arises due to the inherent properties of fingertip skin, such as its elasticity and deformation under pressure~\cite{ross2004estimating,maltoni2009handbook}. As a contactless solution, it offers the added benefit of enhanced hygiene, reducing the risk of spreading infectious diseases like COVID-19. While 3D fingerprint recognition and identification are relatively new advancements, they face significant challenges, particularly in accurately distinguishing ridge and valley patterns in 3D space which is essential for reliable fingerprint recognition.

There are two primary methods for 3D fingerprint recognition and identification.

\subsection{3D Fingerprint Recognition by Unwrapping 3D Fingerprint}

In this method, a 3D fingerprint is converted into a flat, unrolled fingerprint image, which can then be processed using traditional 2D recognition systems. This technique allows for the integration of 3D fingerprint data into existing 2D fingerprint matching frameworks, using the advantages of 3D capture while maintaining compatibility with 2D recognition algorithms.

3D fingerprints can be unrolled into 2D by approximating the finger's surface as a tube~\cite{abramovich2010mobile,wang2010data}, cylinder~\cite{chen20063d}, or sphere~\cite{wang2010fit}. In this process, the finger's surface is divided into slices, with each slice unrolled as a portion of the assumed shape. Previous methods, such as those in \cite{abramovich2010mobile} and \cite{wang2010data}, used Euclidean distances between adjacent points to unroll the 3D fingerprints. Zhao et al. \cite{zhao20113d} proposed a method that introduced a distortion effect during the unrolling process to improve the matching accuracy between 3D and 2D fingerprints. However, these methods have limitations, such as relying on albedo images for pixel intensity assignment, which is not always reliable or available with all 3D scanners. Consequently, having recognition methods based solely on the 3D point cloud, without relying on additional resources like albedo images, would be more desirable.

\subsection{3D Fingerprint Recognition by Using 3D Point Cloud}
Shafaei~\cite{shafaei20113d} utilized surface curvature analysis to identify ridges and valleys in 3D point clouds, enabling the detection of 3D minutiae points. This was among the earliest methods relying solely on 3D point clouds for fingerprint recognition. However, it faced limitations due to numerous false minutiae detections, which significantly impacted recognition performance. Kumar et al.~\cite{kumar2013towards,journals/pami/KumarK15} introduced a photometric stereo-based approach for generating 3D fingerprints from a single camera and developed a 3D minutiae representation technique that outperformed traditional contactless 2D fingerprint identification methods. Labati et al.\cite{labati2015toward} created a 3D model of a finger's shape by capturing images of a moving finger, achieving satisfactory results with fewer constraints compared to touch-based systems. Lin et al.\cite{lin2017tetrahedron} proposed a faster 3D fingerprint identification method using a single camera and multiple LED lights to capture 3D fingerprints efficiently. Yin et al.~\cite{yin20193d} utilized the topology of ridge and valley curves to project them into multiple 2D planes, extracting features from these planes for 3D fingerprint recognition.

This paper introduces a method for using 3D fingerprint point clouds for recognition and identification, divided into three main stages. In the first stage, global fluctuations on the surface of the 3D fingerprint are reduced. In the second stage, the 3D point cloud is unwrapped in $X$ (fingerprint width) and $Y$ (fingerprint length) directions. Lastly, the unwrapped point cloud is converted to gray scale image enabling its use in conventional 2D fingerprint recognition and identification systems.

The main contributions of this paper are as follows:

\begin{enumerate}
    \item Unlike the existing 3D fingerprint unwrapping approaches, this paper introduces a new 3D fingerprint recognition and identification method based on unwrapping the 3D point cloud where the albedo image is not required.
    \item The proposed method is able to reduce the registration issue in 3D fingerprint recognition and identification by unwrapping the 3D point cloud in its length and width directions as well as elimination of global fluctuation while preserving the ridge height and valley depth information.
\end{enumerate}

The rest of this paper is structured as follows: Section~\ref{sec:methodology} describes our proposed method and briefly reviews B-spline curve fitting technique for flatten 3D fingerprint point cloud to generate a grayscale image. Section~\ref{sec:ExperimentalEvaluation} details the experimental setup as well as the results. Section~\ref{sec:Discution} analysis the results obtained, and conclusions are drawn in Section~\ref{sec:Conclusion}.

\section{METHODOLOGY}\label{sec:methodology}
\begin{figure}[!t]
\begin{center}
\begin{center}
\includegraphics[width=0.44\textwidth]{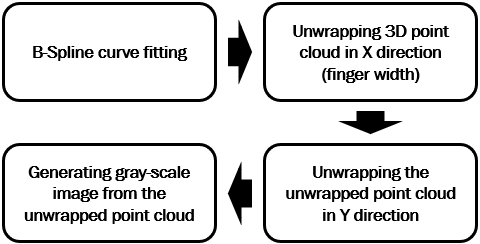}\\
\end{center}
\caption{Flow of processes used in the proposed method.}
\label{fig::flow}
\end{center}
\end{figure}

This section outlines the proposed method for 3D fingerprint recognition by unwrapping 3D point cloud. This method is built on the concept used in~\cite{askarinb}. Flattening the point cloud will reduce or eliminate the impact of topological height, allowing height variations in the flattened image to represent ridge and valley patterns. In this flattened point cloud, the highest points are expected to correspond to ridges, while the lowest points likely indicate valleys within the point cloud. In our approach we use B-Spline curve fitting to unwrap 3D point cloud instead of flattening it. Similarly, 2D images will be generated from the unwrapped 3D point cloud, which can then serve as input for existing fingerprint recognition tools, such as VeriFinger~\cite{VeriFingerSDK}. The flow of the processes is illustrated in Figure~\ref{fig::flow}.

\subsection{B-Spline Curve Fitting}\label{subsec:B-Spline_Curve_Fitting}

In this section B-spline~\cite{komzsik2017approximation} will be introduced. 

\subsubsection{B-spline}

A B-spline curve of degree $k$ can be represented by:
\begin{equation}
\label{B_Spline_curve}
C(t) = \sum_{i=0}^{n} P_i N_{i,k}(t) ,
\end{equation} 

where $N_{i,k}(t)$ are B-spline basis functions and $P_i$ are B-spline control points, and $n$ is the number of control points needed for calculating the basis function, defined as:
\begin{equation}
\label{Control_Points}
n = k + s ,
\end{equation}

where $k$ is the B-spline curve degree, and  $s$ is the number of segments.

The B-spline basis, $N_{i,k}(t)$, is defined recursively by the Cox-de Boor recursion formulae~\cite{cox1972numerical,de1972calculating}.

For $k = 0$:
\begin{equation}
N_{i,k}(t) = \begin{cases} 1 & \text{if } t_i \leq t < t_{i+1} \\ 0 & \text{otherwise} \end{cases}
\end{equation}

For $k > 0$:
\begin{equation}
 N_{i,k}(t) = \frac{t - t_i}{t_{i+k} - t_i} N_{i,k-1}(t) + \frac{t_{i+k+1} - t}{t_{i+k+1} - t_{i+1}} N_{i+1,k-1}(t)
\end{equation}

The control points are found by solving:
\begin{equation}
NP = C ,
\end{equation}

where $P$ is the vector of control points, and $C$ is the vector of data points. The number of knots in the knot vector ($m$) is:
\begin{equation}
m = n + k + 1 .
\end{equation}

The knot vector is a non-decreasing sequence of real numbers that defines segment boundaries:
\begin{equation}
 t = \{t_0, t_1, t_2,...,t_m\},  t_i <=t_{i+1} .
\end{equation}

For a B-spline of degree $k$, The first $k$ elements in the knot vector are equal to the start point of the first segment, and the last $k$ elements are equal to the end point of the last segment.

\subsubsection{Applying B-spline on 3D Point Cloud}
\begin{figure}[!t]
\begin{center}
\begin{minipage}{0.489\textwidth}
\begin{center}
\includegraphics[width=1\textwidth]{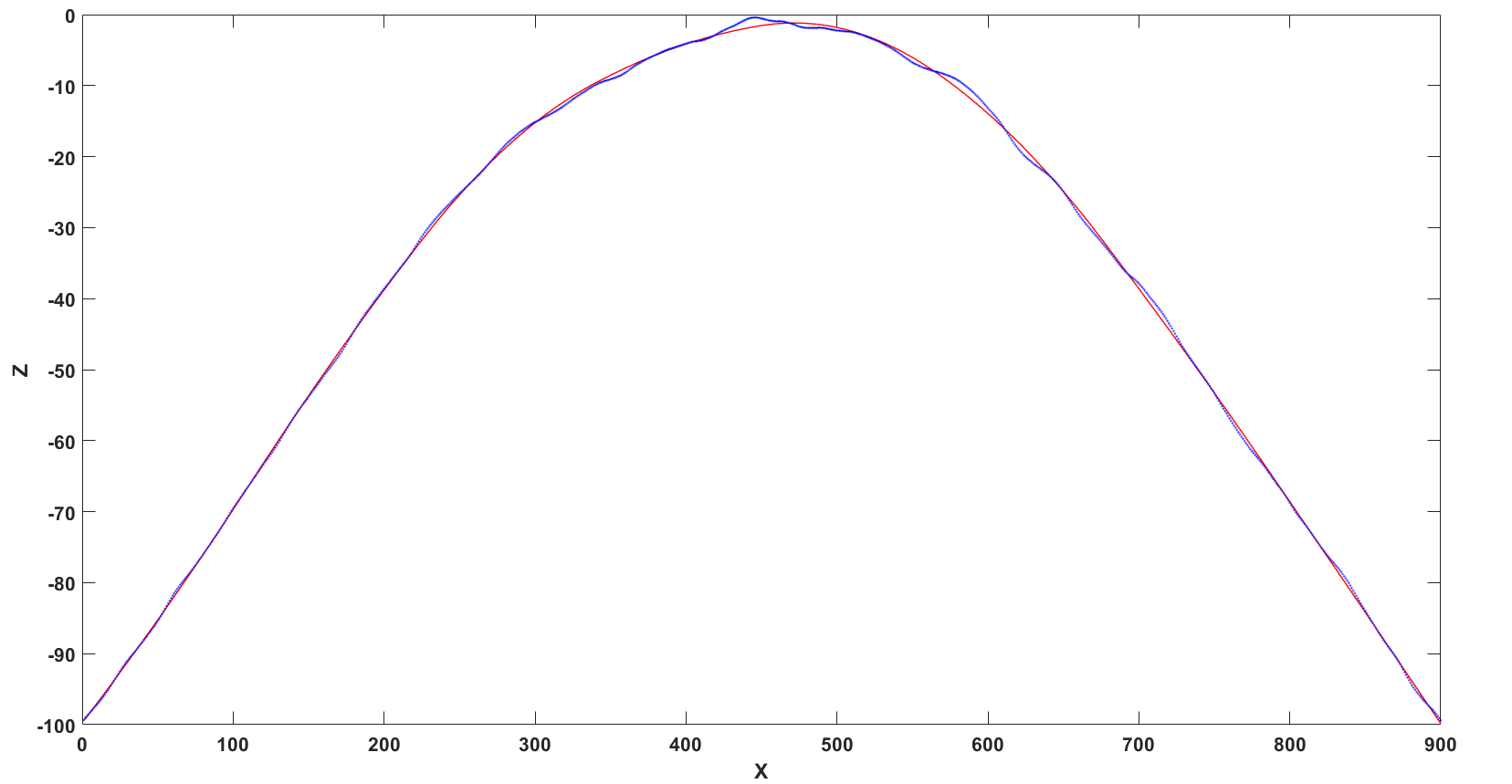}\\ 
(a) 
\end{center}
\end{minipage} \hfil
\begin{minipage}{0.489\textwidth}
\begin{center}
\includegraphics[width=1\textwidth]{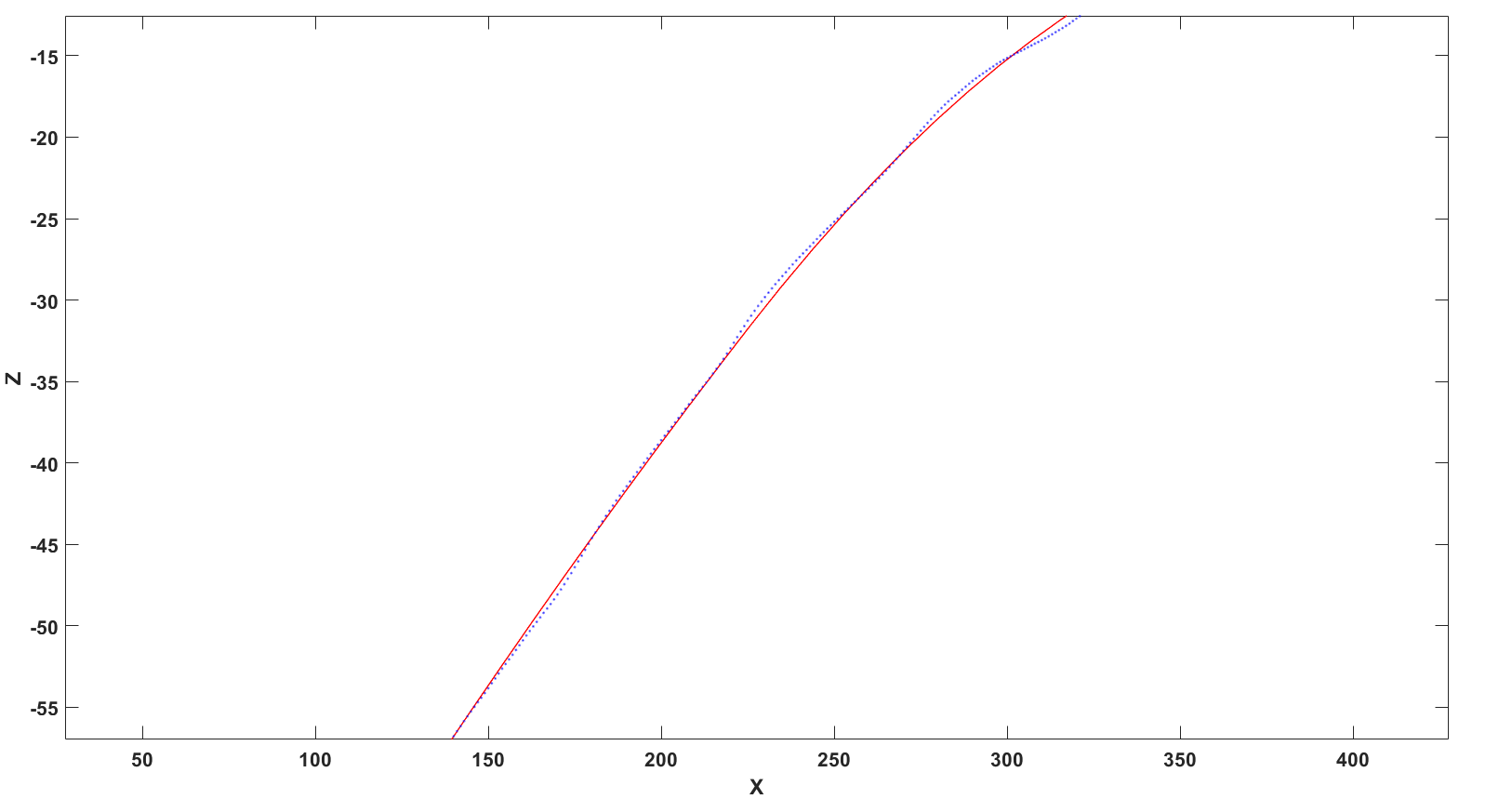}\\
(b) 
\end{center}
\end{minipage}
\caption{(a) shows a sample slice of points in blue and the fitted curve in red. (b) shows a magnified image from a section of (a).}
\label{fig::8PartitionPoly2_B_Spline_Slice}
\end{center}
\end{figure}

Here, we cut point cloud into slices with the width of 1 point in the direction of finger length ($y$ direction) and we apply B-spline curve fitting with 8 partitions and a polynomial degree of 2 as in~\cite{askarinb} Section II F to determine the best-fitting curve for each slice. Figure~\ref{fig::8PartitionPoly2_B_Spline_Slice} (a) shows a sample slice of points in blue and the fitted curve in red. Figure~\ref{fig::8PartitionPoly2_B_Spline_Slice} (b) shows a magnified image from a section of Figure~\ref{fig::8PartitionPoly2_B_Spline_Slice} (a). The points above and below the fitted red curve represent ridges and valleys respectively. Each point’s $Z$ value on the fitted curve is subtracted from the corresponding point’s $Z$ value in the slice to eliminate global fluctuations within each slice. This process will be repeated for all the points. The obtained value for each point will be used as a new $Z$ value for that specific point. The next step to unwrap point cloud is to find updated $X$ and $Y$ values which will be detailed in the next two sections.

\subsection{Unwrapping the 3D Point Cloud in the X Direction (Finger Width)}\label{subsec:Unwrap_X}
\begin{figure}[!t]
\begin{center}
\begin{center}
\includegraphics[width=0.35\textwidth]{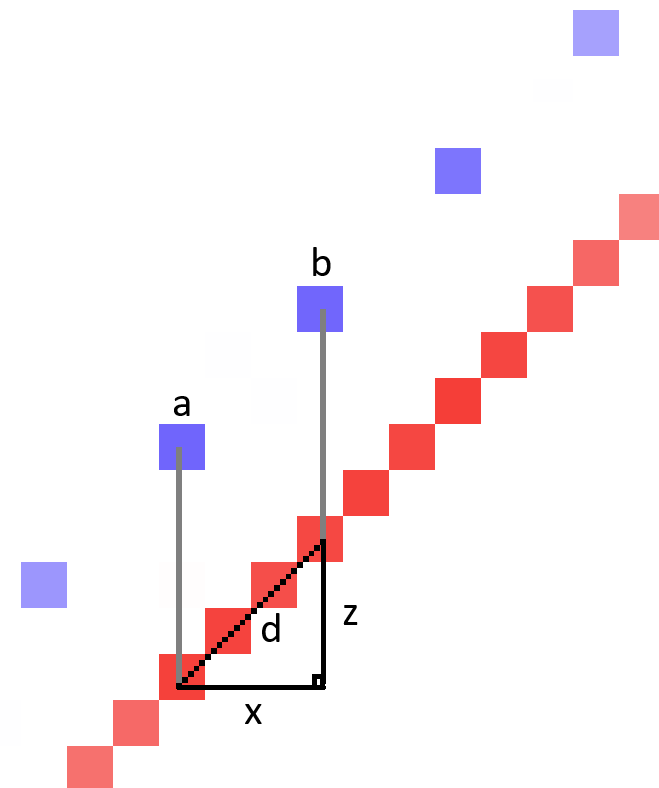}\\
\end{center}
\caption{A magnified section of Figure~\ref{fig::8PartitionPoly2_B_Spline_Slice} (b), showing two adjacent points, a and b. $X$ is the horizontal distance between points a and b, and $Z$ is the difference between the $z$ value of a and b, and d is the distance between the points.}
\label{fig::Magnified_Points_n_Curve}
\end{center}
\end{figure}

To unwrap the point cloud in the $X$ direction, each slice will be unwrapped individually. The distance between any adjacent points in a slice can be estimated by calculating the distance between their corresponding points on the fitted curve. Figure~\ref{fig::Magnified_Points_n_Curve} is a magnified section of Figure~\ref{fig::8PartitionPoly2_B_Spline_Slice} (b), showing two adjacent points, a and b. $X$ is the horizontal distance between points a and b, and Z is the difference between the $z$ value of a and b which both can be calculated from the coordinates of a and b in the point cloud. By having X and Z, d, the estimated distance of a and b can be calculated by solving:

\begin{equation}\label{eq_triangle}
 d = \sqrt{X^2 + Z^2} .
\end{equation}

d for all the points in a slice can be calculated and will be used as a new distance between points in a slice. This process will be repeated for all the slices to obtain the $X$ values for the unwrapped point cloud. 

\subsection{Unwrapping the X-Direction Unwrapped Point Cloud in the Y Direction}\label{subsec:Unwrap_Y}
\begin{figure}[!t]
\begin{center}
\begin{minipage}{0.489\textwidth}
\begin{center}
\includegraphics[width=1\textwidth]{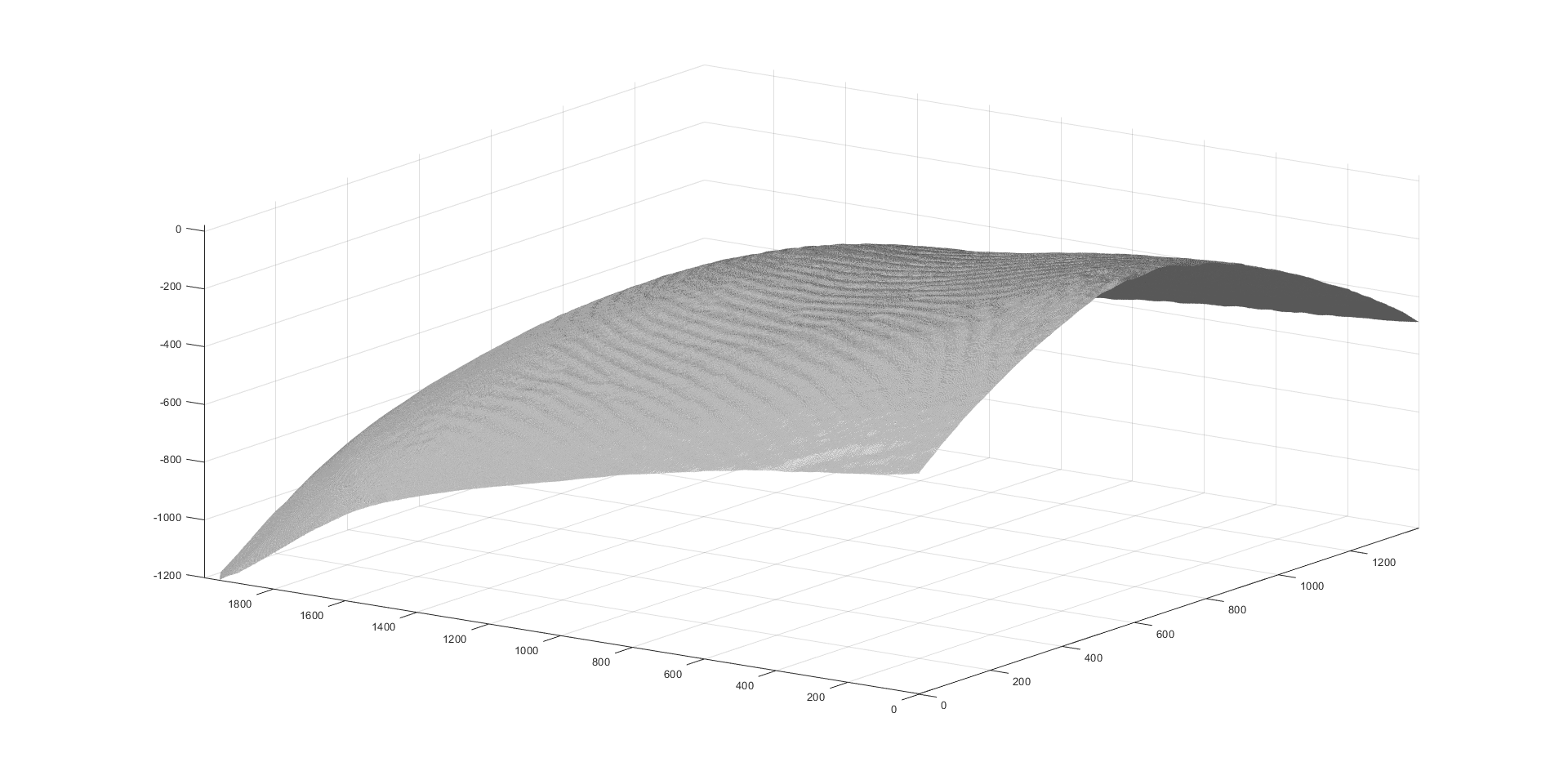}\\ 
(a) DB1-14-4-3D
\end{center}
\end{minipage} \hfil
\begin{minipage}{0.489\textwidth}
\begin{center}
\includegraphics[width=1\textwidth]{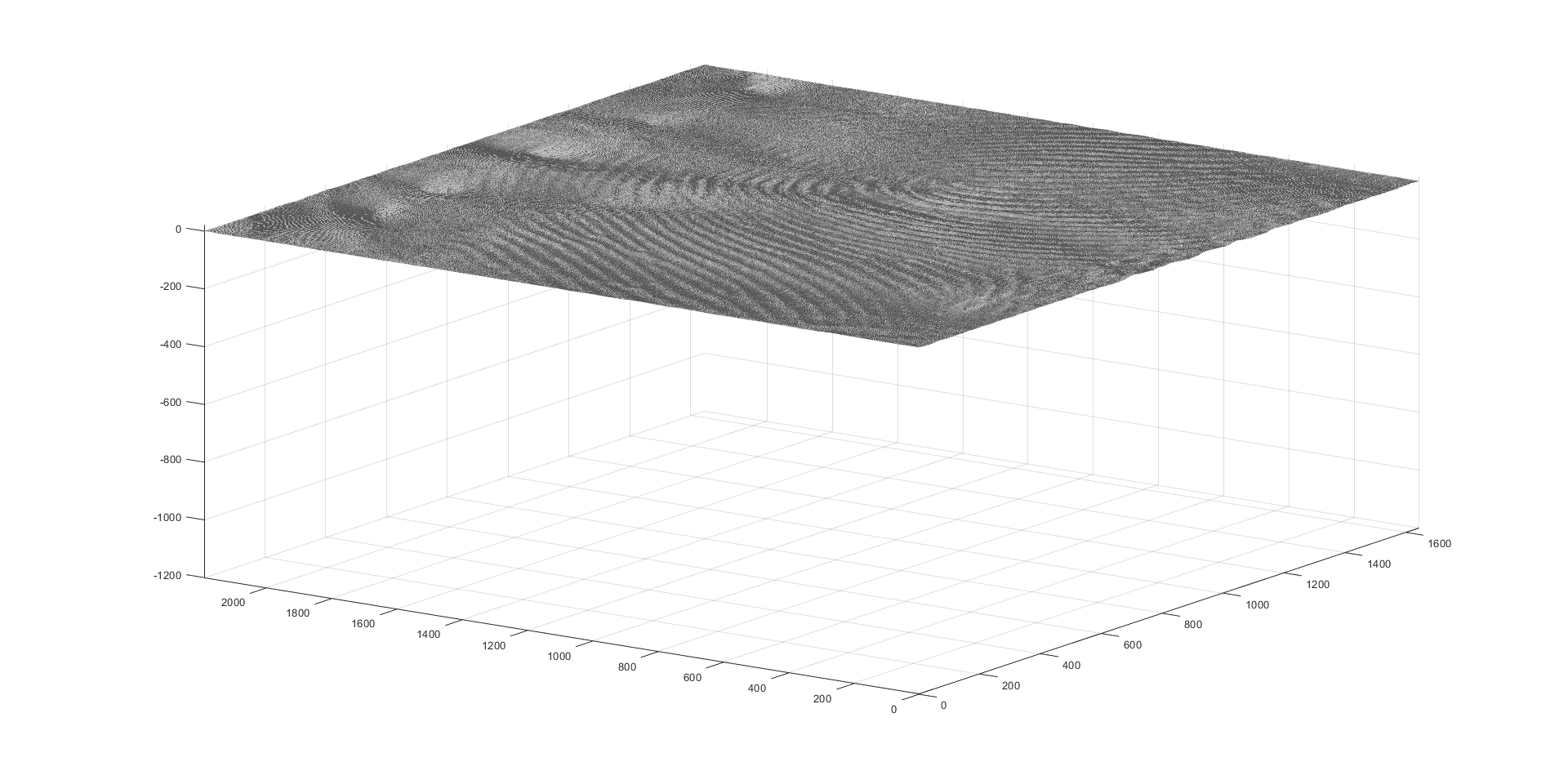}\\
(b) DB1-14-4-Unwrapped
\end{center}
\end{minipage}
\begin{minipage}{0.489\textwidth}
\begin{center}
\includegraphics[width=1\textwidth]{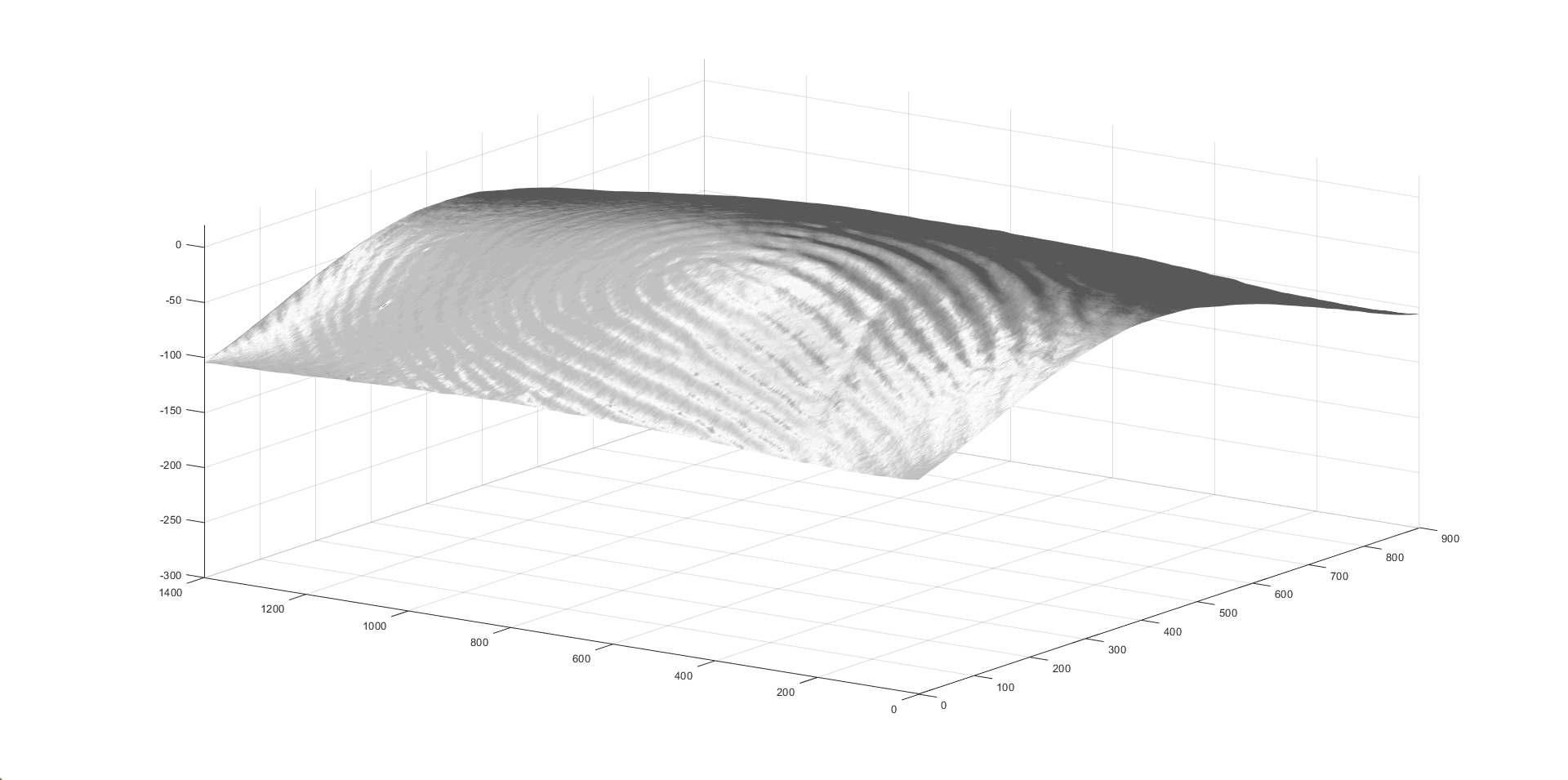}\\
(c) DB2-S2-6-2-3D
\end{center}
\end{minipage}
\begin{minipage}{0.489\textwidth}
\begin{center}
\includegraphics[width=1\textwidth]{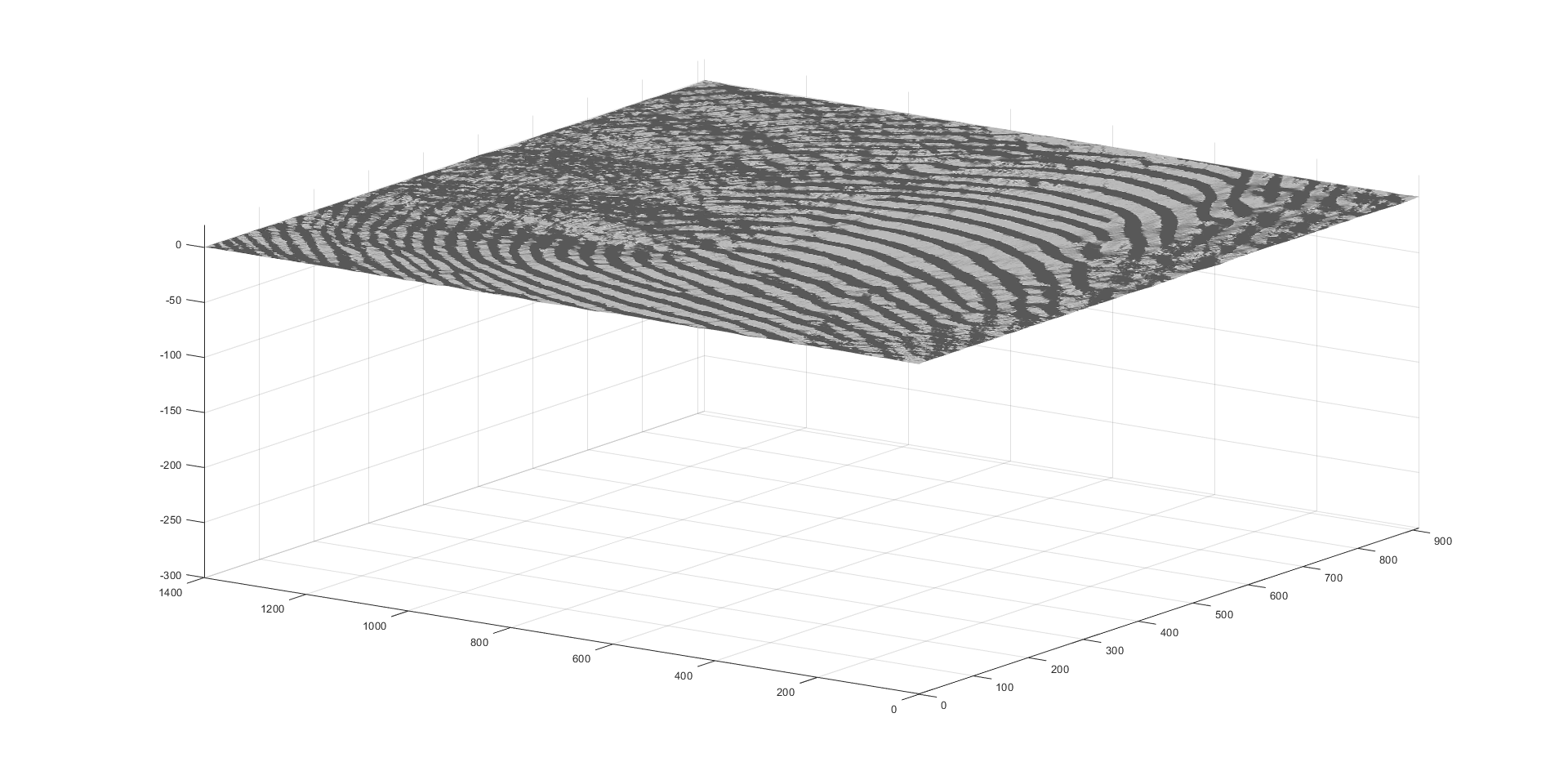}\\
(d) DB2-S2-6-2-Unwrapped
\end{center}
\end{minipage}
\caption{(a) and (c) show the 3D point clouds of impressions 14-4 from Database 1 and 6-2 from Database 2, Session 2, respectively, plotted using Matlab. (b) and (d) display their corresponding unwrapped 3D point clouds in both the $X$ and $Y$ directions using the proposed method.}
\label{fig::3DVSUnwrapped_DB1_DB2_S2}
\end{center}
\end{figure}

Similar to the unwrapping process in $X$ direction, the unwrapped point cloud in $X$ direction can be unwrapped in $Y$ direction. First, the unwrapped point cloud in $X$ direction will be cut into slices with the width of 1 point in the direction of finger width ($X$ direction). For every adjacent two points on the slice, there are two adjacent fitted curves from the previous section (\ref{subsec:Unwrap_X}). The corresponding points on these curves to the points on the unwrapped point cloud can be found by using $X$ and $Y$ coordinates of the points in the unwrapped point cloud. By using the detected points on the curves, the distance ($d$) between the points can be calculated with~\ref{eq_triangle}. The distance $d$ for each point within a slice can be computed and then applied as a new spacing measure between points in that slice. This process is repeated for all slices to determine the Y-coordinates for the unwrapped point cloud. Figure~\ref{fig::3DVSUnwrapped_DB1_DB2_S2} (a) and (c) show the 3D point clouds of impressions 14-4 from Database 1 and 6-2 from Database 2, Session 2, respectively, plotted using Matlab. Figure~\ref{fig::3DVSUnwrapped_DB1_DB2_S2} (b) and (d) display their corresponding unwrapped 3D point clouds in both the $X$ and $Y$ directions using the proposed method.

\subsection{Generating Gray-Scale Image From the Unwrapped Point Cloud}\label{subsec:To_Gray}
\begin{figure}[!t]
\begin{center}
\begin{minipage}{0.24\textwidth}
\begin{center}
\includegraphics[width=1\textwidth]{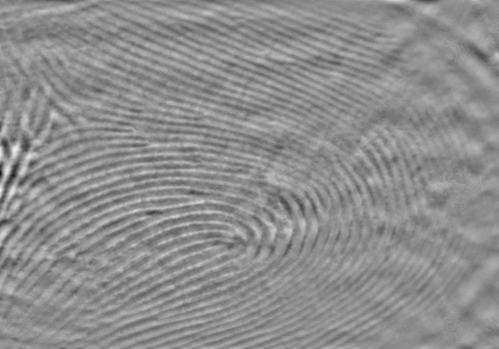}\\ 
(a) DB1-14-4-Flattened
\end{center}
\end{minipage} \hfil
\begin{minipage}{0.24\textwidth}
\begin{center}
\includegraphics[width=1\textwidth]{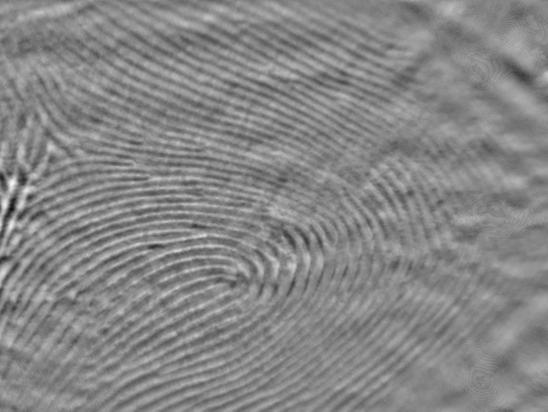}\\
(b) DB1-14-4-Unwrapped
\end{center}
\end{minipage}
\begin{minipage}{0.24\textwidth}
\begin{center}
\includegraphics[width=1\textwidth]{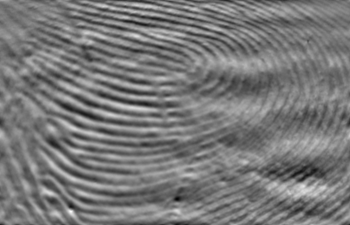}\\
(c) DB2-S2-6-2-Flattened
\end{center}
\end{minipage}
\begin{minipage}{0.24\textwidth}
\begin{center}
\includegraphics[width=1\textwidth]{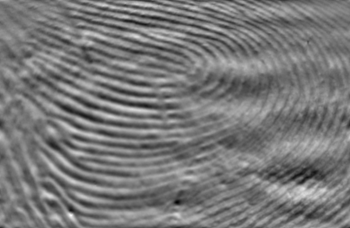}\\
(d) DB2-S2-6-2-Unwrapped
\end{center}
\end{minipage}
\caption{(a) shows the generated gray-scale image from the flattened point cloud of impression 14-4 from Database 1, and (c) presents the gray-scale image from the flattened point cloud of impression 6-2 from Database 2, Session 2, both generated using~\cite{askarinb}. (b) and (d) display the grayscale images generated from the unwrapped 3D point clouds in Figure~\ref{fig::3DVSUnwrapped_DB1_DB2_S2} (b) and (d) respectively, using the proposed method.}
\label{fig::FlattenedVSUnwrapped_DB1_DB2_S2}
\end{center}
\end{figure}

With a similar method used in~\cite{askarinb}, to create a grayscale image from a 3D point cloud, the $X$ and $Y$ coordinates of the unwrapped point cloud can be used to represent pixel locations in the fingerprint image, while the heights of the points ($Z$ values of the unwrapped point cloud) can be used to determine pixel intensities. The highest point in the point cloud is mapped to the lowest intensity (i.e., 0) in the gray-scale image, and the lowest point corresponds to the highest intensity (i.e., 255). Points with heights between these extremes are assigned intensities based on their relative height, creating a smooth gradient that reflects the height variations across the fingerprint surface. Figure~\ref{fig::FlattenedVSUnwrapped_DB1_DB2_S2} (a) shows the generated gray-scale image from the flattened point cloud of impression 14-4 from Database 1, and Figure~\ref{fig::FlattenedVSUnwrapped_DB1_DB2_S2} (c) presents the gray-scale image from the flattened point cloud of impression 6-2 from Database 2, Session 2, both generated using~\cite{askarinb}. Figure~\ref{fig::FlattenedVSUnwrapped_DB1_DB2_S2} (b) and (d) display the gray-scale images generated from the unwrapped 3D point clouds in Figure~\ref{fig::3DVSUnwrapped_DB1_DB2_S2} (b) and (d) respectively, using the proposed method.

\section{Experimental Evaluation}\label{sec:ExperimentalEvaluation}
This section outlines the datasets used and the experiments performed for the evaluation of the proposed method. 


\subsection{Dataset}
Our experiments utilize the Hong Kong Polytechnic University 3D Fingerprint Images Database Versions 1~\cite{HongKongDBv1,kumar2013towards,journals/pami/KumarK15} (Database one) and 2~\cite{HongKongDBv2,lin2017tetrahedron} (Database two). 

Database Version 2 was collected over two sessions. In the first session, data was gathered from 336 individuals, but three impressions (from individuals 165, 174, and 205) were excluded due to corrupted files. To be able to compare our results with existing works, the point clouds from the first 300 remaining individuals are used in our experiments. Each individual has six three-dimensional (3D) fingerprint impressions, represented as 3D point clouds (900 by 1400 points), with each point provided with three coordinates that indicate its location and height.

The second session of this database includes data collected from 200 individuals. The fingerprint impressions for the individual with ID 86 were corrupted and are therefore excluded from our experiments. For comparing the results with the existing tests, the point cloud data from the first 160 remaining individuals are used in our experiments. Point clouds in this session also have the same dimensions as the first session (900 by 1400 points).

Hong Kong Polytechnic University 3D Fingerprint Database version 1~\cite{HongKongDBv1} contains fingerprint data from 260 individuals, with six impressions per person. Each 3D fingerprint impression is represented as a point cloud with dimensions of 1394 by 1994 points. To enable comparison with other studies, our experiments use the fingerprint impressions from the first 240 individuals in this database.

\subsection{Experiments}
This section details the tool setup and assessment process for the proposed approach for 3D fingerprint recognition and identification.

\subsubsection{Tools}
A Dell workstation equipped with a 12th-generation Intel Core i9 processor, 32GB of RAM, and running Microsoft Windows 11 was used for unwrapping 3D point clouds, converting them into grayscale images, and conducting recognition and identification tests. Matlab R2023b was utilized for unwrapping 3D fingerprint point clouds and creating grayscale images~\cite{Matlab}. Visual Studio 2022~\cite{VisualStudio} and the VeriFinger SDK 13.1~\cite{VeriFingerSDK} were employed to generate fingerprint templates and compute matching scores.


\subsubsection{Set up}
The B-Spline curve fitting parameters are set based on~\cite{askarinb}. Specifically, for Database 2 Session 1 and Database 2 Session 2, curve fitting is performed with a polynomial degree of 2 and 8 partitions, while for Database 1, a polynomial degree of 2 and 12 partitions is used.

The selection of these parameters is based on the fact that higher polynomial degrees increase computation time. Therefore, lower polynomial degrees are preferred. In addition, curve fitting with higher polynomial degrees can lead to overfitting. On the other hand, curve fitting with lower polynomial degrees is less accurate for large partitions. As a result, using multiple small partitions is preferred over a single large partition. Since the point clouds in Database 1 are not cropped and have larger dimensions, a greater number of partitions was required to obtain optimal performance.



\subsubsection{Experiment A}\label{sec:ExperimentA}
In our first experiment, fingerprints of the first session of Database 2 are used. After removing the impressions from the subjects with corrupted data, the impressions from the first 300 subjects are used. Next, by applying the proposed method, the 3D point clouds were unwrapped and converted to gray-scale images. Then VeriFinger is used to calculate the matching scores. To be able to compare our results with existing works, the protocol in~\cite{lin2017tetrahedron} is used which uses the imposter matching scores for all the 6 impressions of all the subjects. in total, 4500 genuine and 1614600 imposter matching scores were calculated. 

\begin{figure}[!t]
\begin{center}
\begin{minipage}{0.489\textwidth}
\begin{center}
\includegraphics[width=1\textwidth]{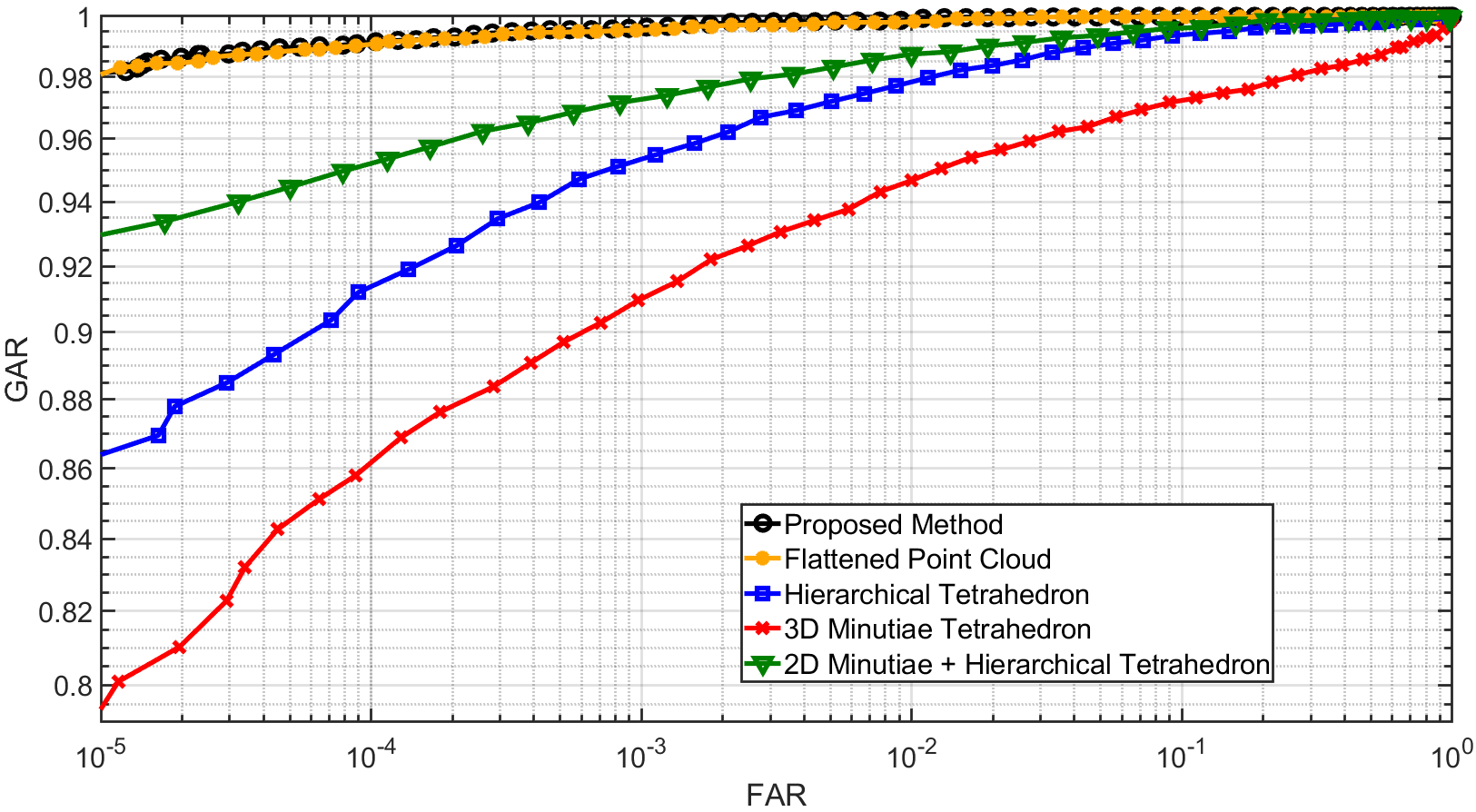}\\ 
(a) ROC
\end{center}
\end{minipage} \hfil
\begin{minipage}{0.489\textwidth}
\begin{center}
\includegraphics[width=1\textwidth]{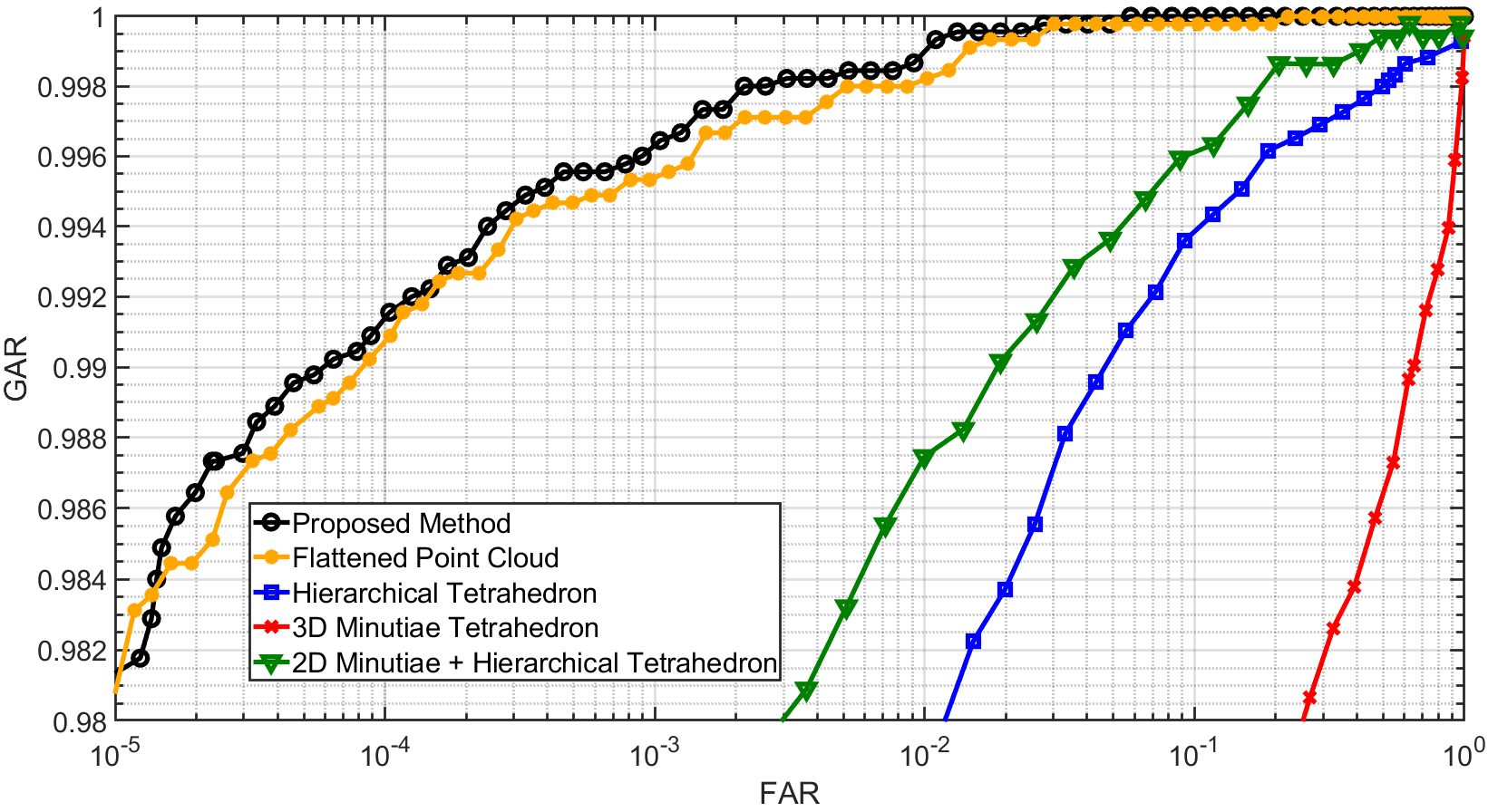}\\
(b) ROC magnified
\end{center}
\end{minipage}
\caption{(a) and (b) show the ROC curve of the proposed method for session 1 of database 2 compared to the conventional methods.}
\label{fig::ROCS1}
\end{center}
\end{figure}

\begin{figure}[!t]
\begin{center}
\begin{center}
\includegraphics[width=0.489\textwidth]{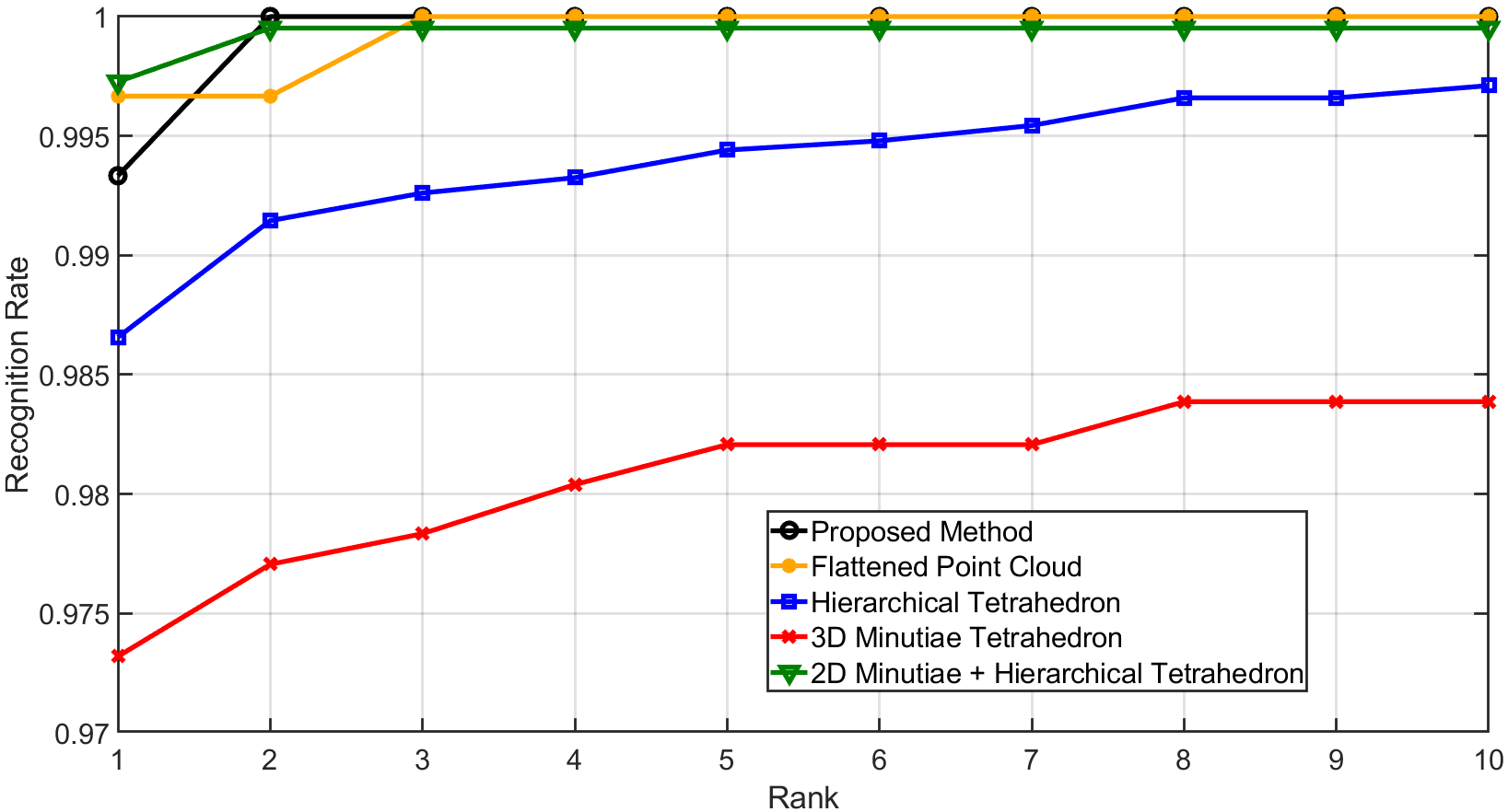}\\
\end{center}
\caption{CMC curve of the proposed method for session 1 of database 2 compared to the conventional methods.}
\label{fig::CMCS1}
\end{center}
\end{figure}

Figure~\ref{fig::ROCS1} and Figure~\ref{fig::CMCS1} display the Receiver Operating Characteristic (ROC) and Cumulative Match Characteristic (CMC) curves for both the proposed and existing methods respectively. The CMC curve is determined by identifying each subject's first impression within a gallery that includes the subject's second impression and all impressions from 299 other subjects, resulting in a gallery size of 1795 impressions. Table~\ref{table:EERComparison} presents a comparison of EERs and rank-1 accuracies. The EER of the proposed method compared to the flattened point cloud was 40\% improved. On the other hand, the rank-1 accuracy of the proposed method was slightly lower than the flattened point cloud method as two subjects failed to be identified as rank-1 in contrast to the one failed rank-1 identified subject in the flattened point cloud but, the rank-2 accuracy of the unwrapped method achieved 100\% accuracy which was higher than flattening method. The rank-1 accuracy result reported in~\cite{lin2017tetrahedron} reflects a test with over 300 subjects. however, the exact number of subjects used is unspecified. Therefore, a direct comparison with our rank-1 accuracy for evaluation is not feasible.

 

\begin{table}[!ht]
\begin{center}
\caption{\label{table:EERComparison}Comparison of EER of the proposed method for session 1 of database 2 with existing methods.}
\resizebox{0.4893\textwidth}{!}{
\begin{tabular}{|c|c|c|}
\hline
\textbf{Experiments}&\textbf{EER}&\textbf{Rank-1}\\

&&\textbf{accuracy}\\
\hline
3D Minutiae Tetrahedron~\cite{lin2017tetrahedron} & 3.50\%&97.23\%\\ 
\hline
3D Hierarchical Tetrahedron~\cite{lin2017tetrahedron} & 1.52\%&98.61\%\\ 
\hline
2D Minutiae + & &\\

3D Hierarchical Tetrahedron~\cite{lin2017tetrahedron}& 1.41\%&99.72\%\\
\hline 
Flattened Point Cloud~\cite{askarinb} &  0.2974\%&99.67\%\\
\hline 
Proposed method &  0.2072\%&99.33\%\\
\hline
\end{tabular}
}
\end{center}
\end{table}

\subsubsection{Experiment B}

In the second experiment, fingerprints from the second session of Database 2 were used. The subjects with corrupted files were removed and the first 160 subjects were selected. The 3D point clouds were unwrapped and converted to grayscale images using the propped method. Following the same protocol as in Experiment A, imposter matching scores were calculated for all six 3D fingerprint impressions of each subject, resulting in 2400 genuine and 457,920 imposter scores. 

\begin{figure}[!t]
\begin{center}
\begin{minipage}{0.489\textwidth}
\begin{center}
\includegraphics[width=1\textwidth]{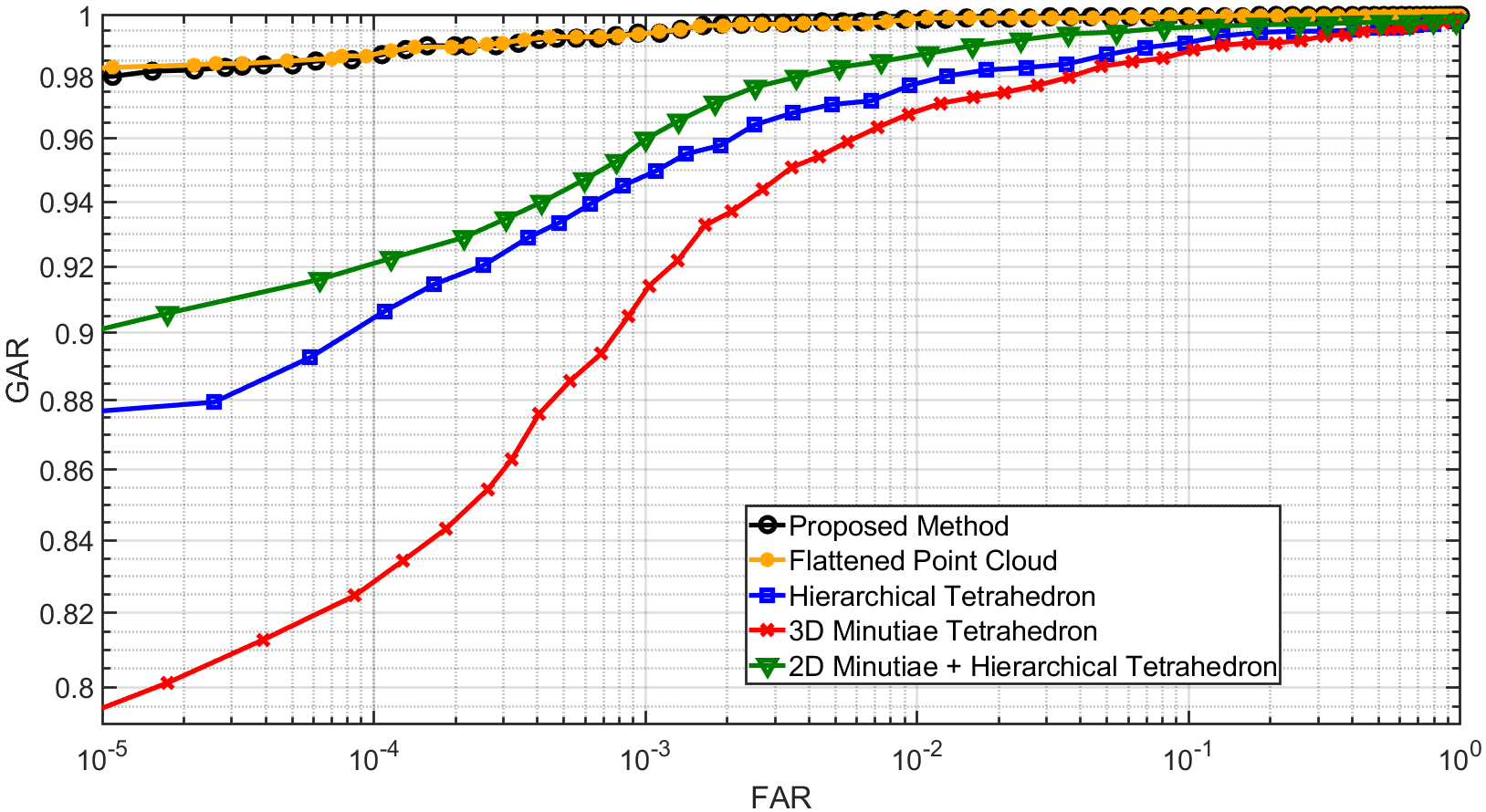}\\ 
(a) ROC
\end{center}
\end{minipage} \hfil
\begin{minipage}{0.489\textwidth}
\begin{center}
\includegraphics[width=1\textwidth]{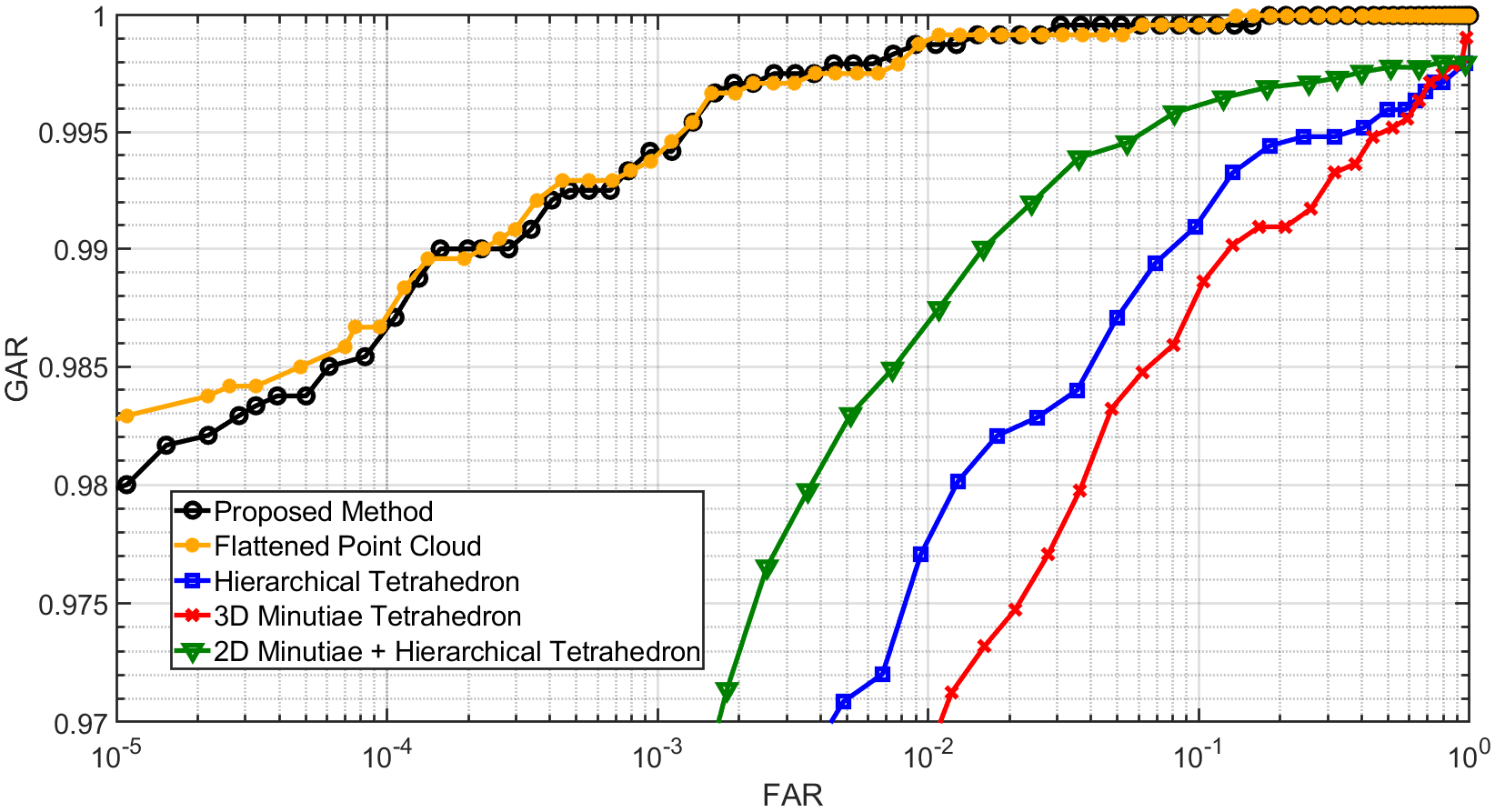}\\
(b) ROC magnified
\end{center}
\end{minipage}
\caption{(a) and (b) show the ROC curve of the proposed method for session 2 of database 2 compared to the conventional methods.}
\label{fig::ROCS2}
\end{center}
\end{figure}

\begin{figure}[!t]
\begin{center}
\begin{center}
\includegraphics[width=0.489\textwidth]{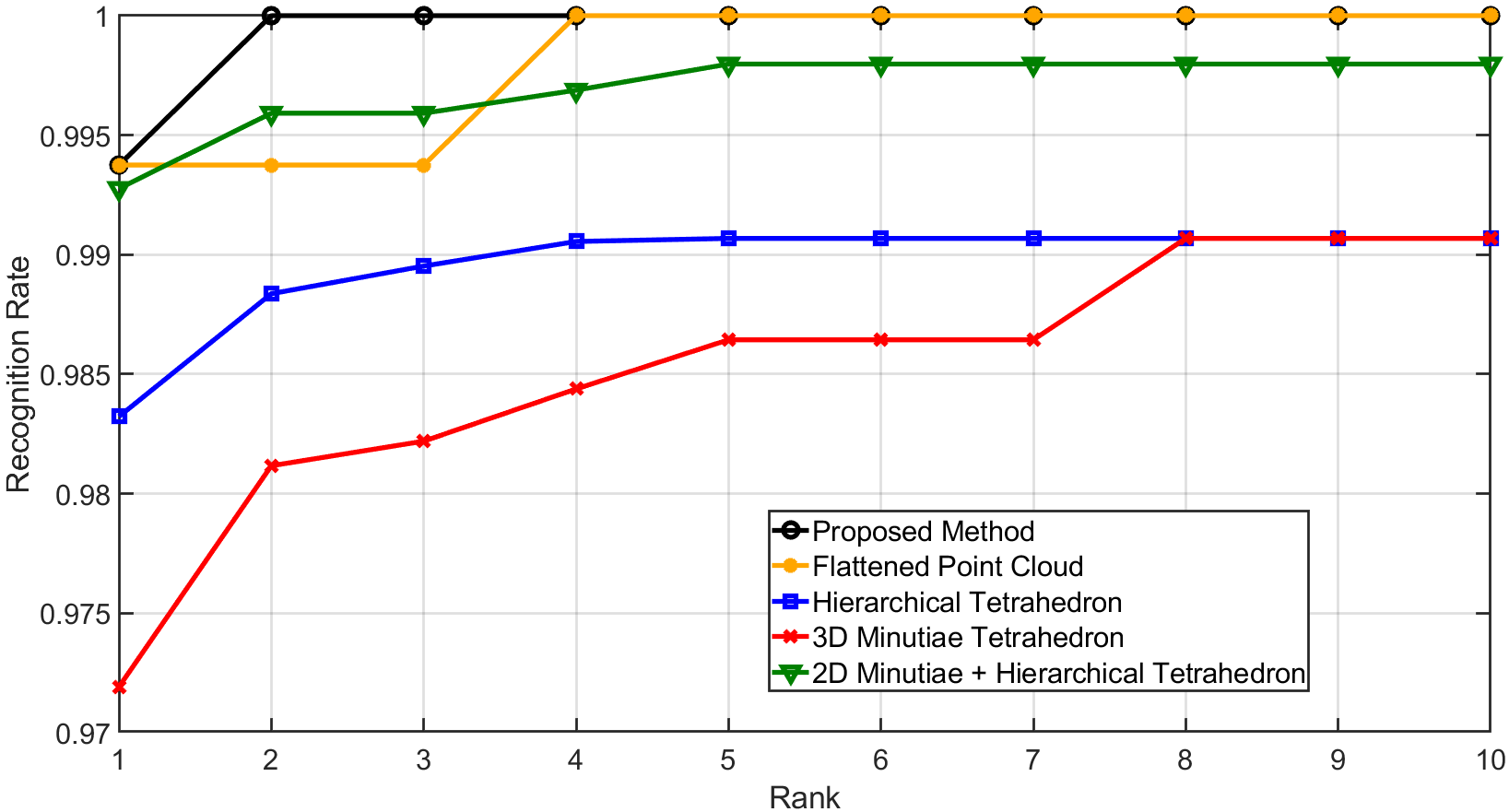}\\
\end{center}
\caption{CMC curve of the proposed method for session 2 of database 2 compared to the conventional methods.}
\label{fig::CMCS2}
\end{center}
\end{figure}

The proposed method’s performance is assessed using ROC and CMC curves, displayed with existing methods in Figure~\ref{fig::ROCS2} and Figure~\ref{fig::CMCS2}, respectively. The CMC curve was determined by matching each subject's first impression in a gallery containing the subject's second impression and all impressions of the other 159 subjects, creating a gallery size of 955 impressions. Table~\ref{table:EERComparisonDBv2_S2} compares the EERs and rank-1 accuracies, showing that the proposed method achieves superior results in both metrics with 0.07\% improvement of EER in comparison to the flattened point cloud method. The 99.38\% rank-1 accuracy is achieved by identifying 159 out of 160 subjects at rank-1 and one subject at rank-2, which is the second-highest possible accuracy after 100\%.


\begin{table}[!ht]
\begin{center}
\caption{\label{table:EERComparisonDBv2_S2}Comparison of EER of the proposed method for session 2 of database 2 with existing methods.}
\resizebox{0.4893\textwidth}{!}{
\begin{tabular}{|c|c|c|}
\hline
\textbf{Experiments}&\textbf{EER}&\textbf{Rank-1}\\
&&\textbf{accuracy}\\
\hline
3D Minutiae Tetrahedron~\cite{lin2017tetrahedron} & 2.41\%&97.29\%\\ 
\hline
3D Hierarchical Tetrahedron~\cite{lin2017tetrahedron} & 1.41\%&98.34\%\\ 
\hline
2D Minutiae + & &\\
3D Hierarchical Tetrahedron~\cite{lin2017tetrahedron}& 1.25\%&99.27\%\\
\hline 
Flattened Point Cloud~\cite{askarinb} &  0.28\%&99.38\%\\
\hline 
Proposed method &  0.26\%&99.38\%\\
\hline
\end{tabular}
}
\end{center}
\end{table}

\subsubsection{Experiment C}
In the third experiment, fingerprints from Database 1~\cite{HongKongDBv1} were used. To compare our results with other methods, only the impressions from the first 240 subjects were selected. The 3D point clouds were unwrapped and converted to grayscale images. To calculate EER, the same protocol as the existing works was used by calculating imposter matching scores for each subject's first 3D fingerprint impression. This resulted in 3,600 genuine (240×6×5/2) and 28,680 (240×239/2) imposter matching scores. 

\begin{figure}[!t]
\begin{center}
\begin{center}
\includegraphics[width=0.489\textwidth]{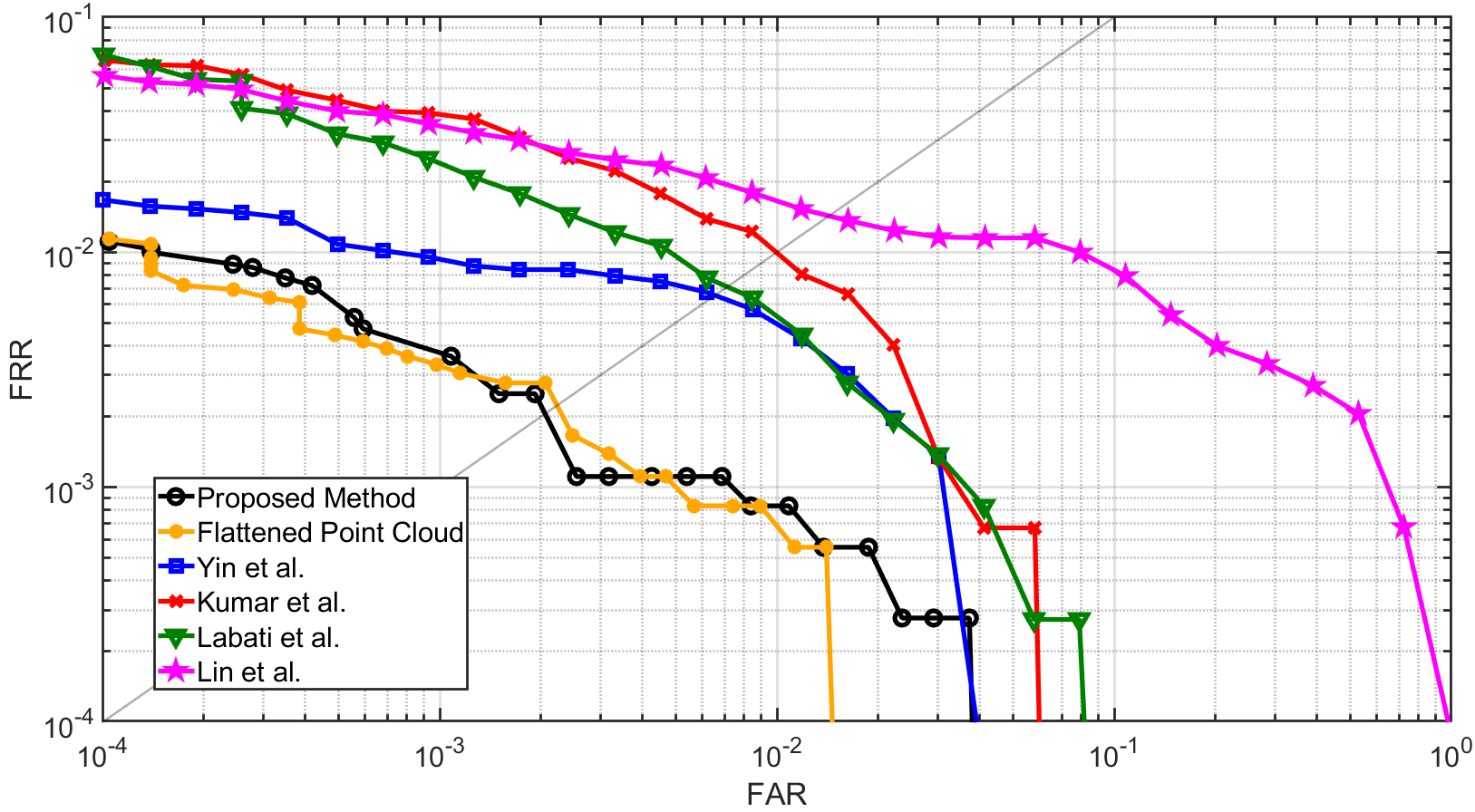}\\
\end{center}
\caption{DET curve of the proposed method on database 1 compared to the conventional methods.}
\label{fig::DET_DB1}
\end{center}
\end{figure}

\begin{figure}[!t]
\begin{center}
\begin{center}
\includegraphics[width=0.489\textwidth]{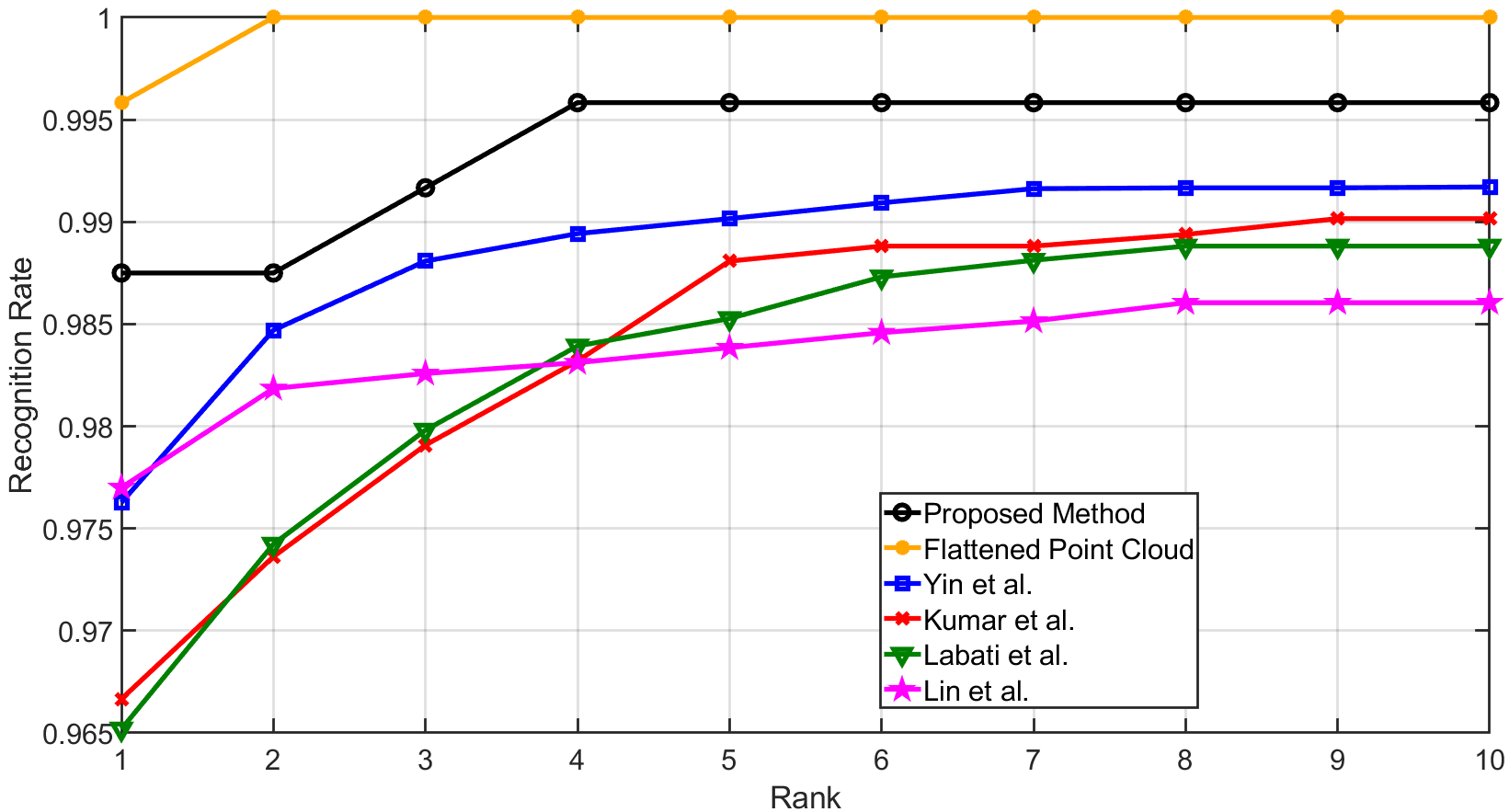}\\
\end{center}
\caption{CMC curve of the proposed method for database 1 compared to the conventional methods.}
\label{fig::CMC_DB1}
\end{center}
\end{figure}

The proposed method's performance on this database is evaluated using Detection Error Trade-off (DET) and CMC curves, shown with the results of existing methods in Figure~\ref{fig::DET_DB1} and Figure~\ref{fig::CMC_DB1} respectively. The CMC curve was obtained by identifying each subject's first impression within a gallery containing the subject's second impression and all impressions of 239 other subjects, creating a gallery size of 1435 impressions. Table~\ref{table:EERComparisonDBv1} compares EERs and rank-1 accuracies. The EER was improved by \%9 compared to the flattened point cloud method but rank-1 accuracy was slightly dropped. However, the proposed method achieved higher EER and Rank-1 accuracy than the rest of the existing methods.

\begin{table}[!ht]
\begin{center}
\caption{\label{table:EERComparisonDBv1}Comparison of EER of the proposed method with existing methods for database 1.}
\resizebox{0.44\textwidth}{!}{
\begin{tabular}{|c|c|c|}
\hline
\textbf{Experiments}&\textbf{EER}&\textbf{Rank-1
accuracy}\\
\hline
Yin et al.~\cite{yin20193d} & 0.68\%&97.6 \%\\ 
\hline
Kumar et al.~\cite{journals/pami/KumarK15} & 1.02\%&96.7\%\\ 
\hline
Labati et al.~\cite{labati2015toward} & 0.70\%&96.5\%\\
\hline 
Lin et al.~\cite{lin2017tetrahedron} & 1.51\%&97.7\%\\
\hline
Flattened Point Cloud~\cite{askarinb} &  0.24\%&99.58\%\\
\hline
Proposed method &  0.22\%&98.75\%\\
\hline
\end{tabular}
}
\end{center}
\end{table}

\subsubsection{Experiment D}
160 subjects in Database 2 Session 1 also provided their fingerprint impressions in Database 2 Session 2. In this experiment, the 3D fingerprints of the 160 mutual subjects of fingerprints of Database 2 session 2 and Database 2 session 1 are used for recognition and identification tests. 

For the genuine matching, the matching score of each impression of a mutual subject in Session 2 with all the impressions of the corresponding subject in Session 1 is calculated, which makes a total of 5760(6x6x160) genuine matching scores. For imposter matching, the matching score of each impression of a subject in Session 2 with all the imposter impressions of the mutual subject in Session 1 is calculated which makes a total of 457,920 imposter matching scores. 

The identification test was performed by identifying the first impression of each subject in Database two Session 2 within a gallery that included the corresponding second impression from Database 2 Session 1 along with all impressions from the other mutual 159 subjects in Database 2 Session one, resulting in a gallery size of 955 impressions. This identification test was also performed for all 15 possible combinations of genuine scores of each subject and the average accuracy was calculated. 

\begin{figure}[!t]
\begin{center}
\begin{center}
\includegraphics[width=0.489\textwidth]{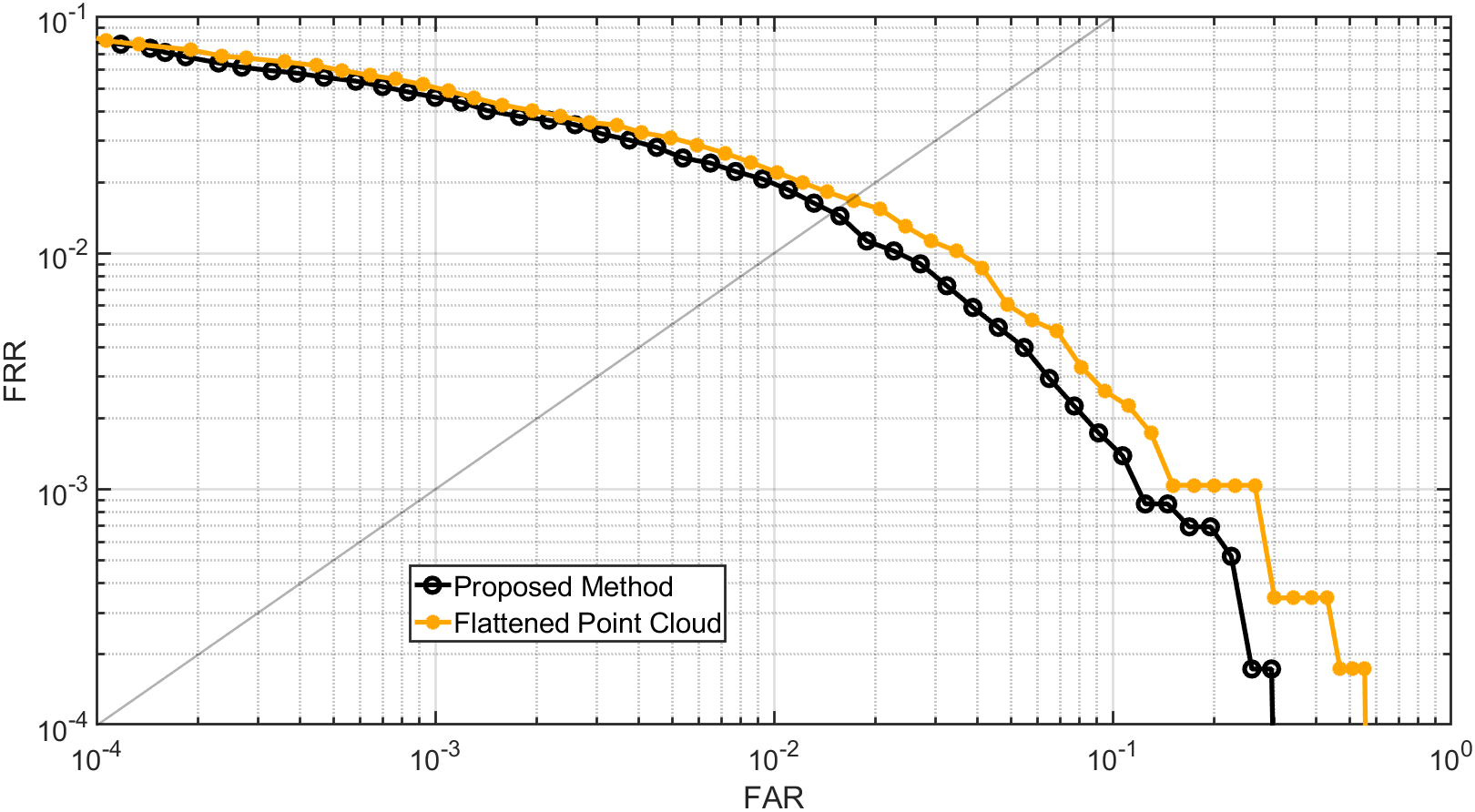}\\
\end{center}
\caption{DET curve of the proposed method for database 2 session 2 vs database 2 session 1 compared to the conventional methods.}
\label{fig::DET_DB2_S2_VS_DB2_S1}
\end{center}
\end{figure}

\begin{figure}[!t]
\begin{center}
\begin{minipage}{0.489\textwidth}
\begin{center}
\includegraphics[width=1\textwidth]{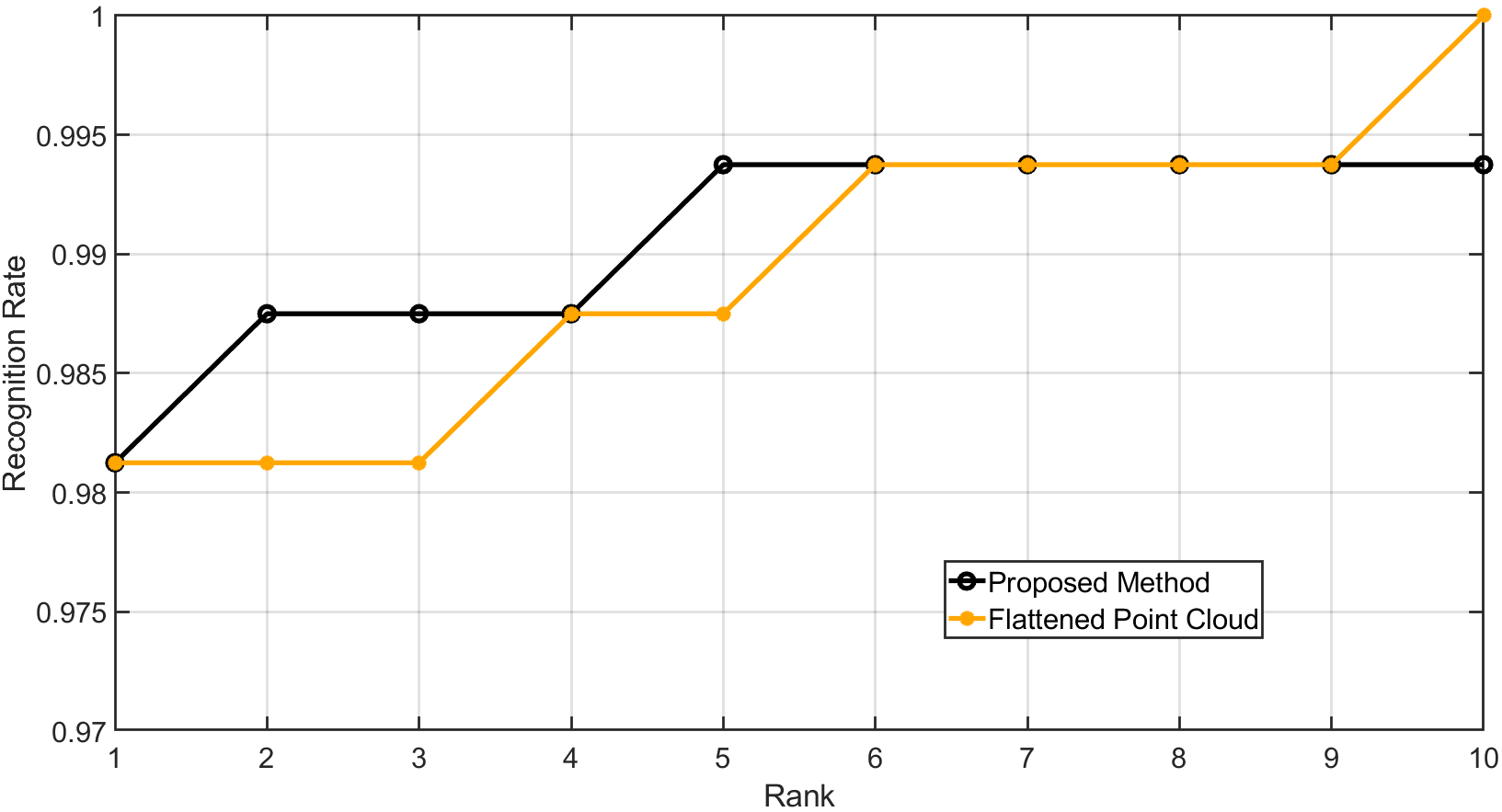}\\ 
(a) CMC
\end{center}
\end{minipage} \hfil
\begin{minipage}{0.489\textwidth}
\begin{center}
\includegraphics[width=1\textwidth]{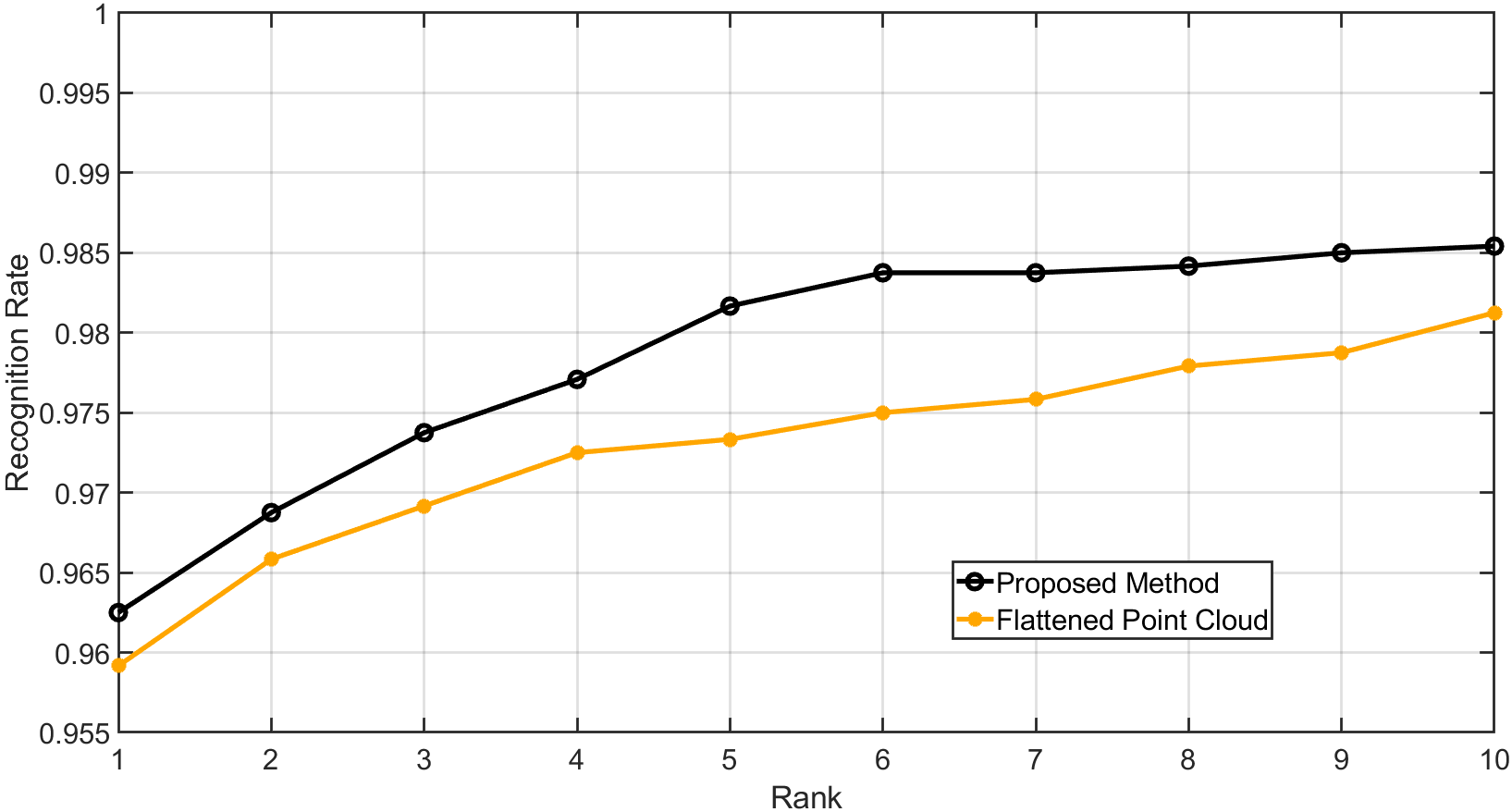}\\
(b) CMC (15 combinations)
\end{center}
\end{minipage}
\caption{CMC curve of the proposed method for database 2 session 2 vs database 2 session 1 compared to the conventional methods.}
\label{fig::CMC_DB2_S2_VS_DB2_S1}
\end{center}
\end{figure}

The DET and CMC curves of the proposed method and the flattened 3D point cloud~\cite{askarinb} are shown in Figure~\ref{fig::DET_DB2_S2_VS_DB2_S1} and Figure~\ref{fig::CMC_DB2_S2_VS_DB2_S1} respectively. The comparison of EERs and rank-1 accuracies are shown in Table~\ref{table:EERComparisonDBv2_S2_VS_DBv2_S1}. The proposed method was able to achieve a better EER than the existing methods. The identification test results also show that the proposed method was able to enhance the identification accuracy in both tests. Table~\ref{table:Time_cost} provides a comparison of the processing time per image in this experiment based on the hardware used. While the proposed method has a higher processing time compared to the flattened point cloud method, it does not introduce delays in practical applications. The time required to generate a grayscale image from the unwrapped 3D point cloud remains shorter than the overall fingerprint acquisition process, so the processing time does not affect real-time usability. Additionally, using more powerful hardware, such as high-performance GPUs and multi-core processors, or adopting cloud computing can further reduce processing time.

\begin{table}[!ht]
\begin{center}
\caption{\label{table:EERComparisonDBv2_S2_VS_DBv2_S1}Comparison of EER and Rank-1 accuracy of the proposed method with flattened point cloud method for database two session two VS database two session one.}
\resizebox{0.489\textwidth}{!}{
\begin{tabular}{|c|c|c|c|}
\hline
\textbf{Experiments}&\textbf{EER}&\textbf{Rank-1}&\textbf{Rank-1 accuracy}\\
&&\textbf{accuracy}&\textbf{15 Combinations}\\
\hline
Flattened Point Cloud~\cite{askarinb} &  1.69\%&98.125\%&95.92\\
\hline
Proposed method &  1.50\%&98.125\%&96.25\\
\hline
\end{tabular}
}
\end{center}
\end{table}

\begin{table}[!ht]
\begin{center}
\caption{\label{table:Time_cost}Comparison of the processing time of the proposed method and the flattened point cloud method based on the hardware used for Database 2, Session 2 versus Session 1.}
\resizebox{0.489\textwidth}{!}{
\begin{tabular}{|c|c|c|}
\hline
\textbf{Experiments}&\textbf{Hardware used}&\textbf{Processing time}\\
&&\textbf{per image (sec)}\\
\hline
Flattened Point Cloud~\cite{askarinb} &  Intel i9, 32GB RAM&0.31\\
\hline
Proposed method &Intel i9, 32GB RAM&2.31\\
\hline
\end{tabular}
}
\end{center}
\end{table}

\subsection{Evaluation and Comparison}
The proposed method had a better recognition performance in all the tests. The identification results of the proposed method on Database 2 were quite similar to that of the Flattened Point Cloud method~\cite{askarinb}. The identification results of the proposed method on Database 1 were slightly lower than the Flattened Point Cloud method~\cite{askarinb} but higher than the other existing methods. The results from the cross-session identification in experiment 4 show that the proposed method was able to get a higher accuracy.

\section{Discussion}\label{sec:Discution}
Even though the unwrapping of a 3D point cloud before conversion to a gray-scale image can cause some distortion and loss of quality, the evaluation of the results from the experiments shows that the proposed method overall can achieve better results than the Flattened Point Cloud method in most of the experiments and obtains a significantly better performance than the rest of the existing methods. This is due to the fact that unwrapping can reduce the registration issue in 3D fingerprint identification.

In the flattening 3D point cloud method, it is essential that during the 3D fingerprint acquisition, the fingers be presented for the camera with the same angle and rotation in order for this method to achieve its best performance. On the other hand, the proposed unwrapping point cloud method does not have this limitation and the fingers can be presented to the camera at any different angle and rotation. This impact is more significant in Experiment D where we used the data that was captured in two different sessions for the recognition and identification tests. 

\begin{figure}[!t]
\begin{center}
\begin{minipage}{0.24\textwidth}
\begin{center}
\includegraphics[width=1\textwidth]{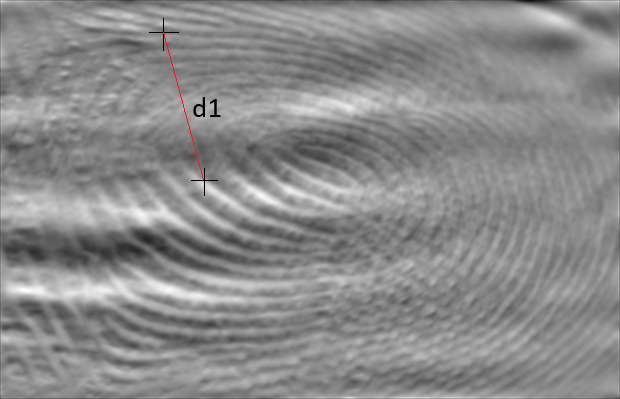}\\ 
(a) DB2-S1-8-5-Flattened
\end{center}
\end{minipage} \hfil
\begin{minipage}{0.24\textwidth}
\begin{center}
\includegraphics[width=1\textwidth]{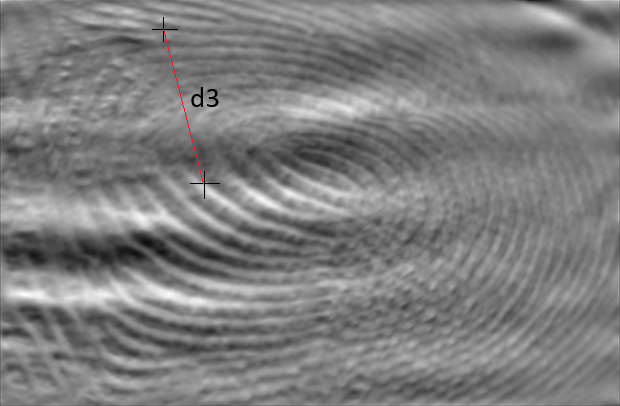}\\
(b) DB2-S1-8-5-Unwrapped
\end{center}
\end{minipage}
\begin{minipage}{0.24\textwidth}
\begin{center}
\includegraphics[width=1\textwidth]{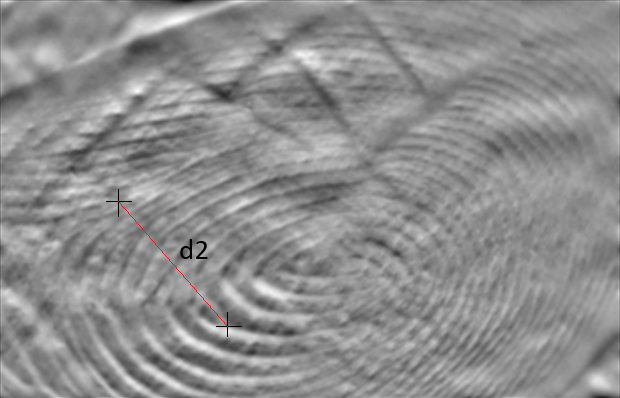}\\
(c) DB2-S2-8-2-Flattened
\vspace{2mm}
\par(a) vs (c)\par
matching score: 42
\end{center}
\end{minipage}
\begin{minipage}{0.24\textwidth}
\begin{center}
\includegraphics[width=1\textwidth]{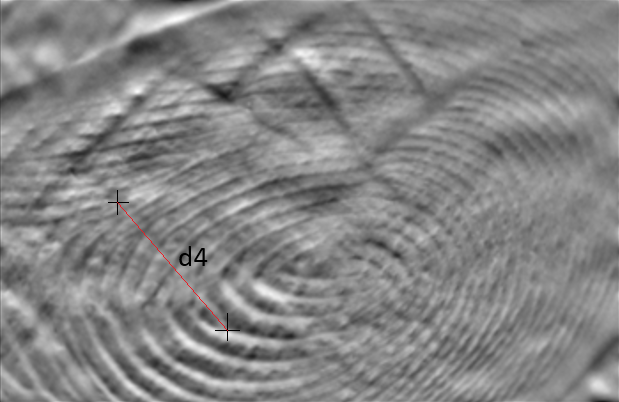}\\
(d) DB2-S2-8-2-Unwrapped
\vspace{2mm}
\par(b) vs (d)\par
matching score: 63
\end{center}
\end{minipage}
\caption{(a) shows flattened impression 8-5 from Database 2 Session 1 and (c) shows flattened impression 8-2 from Database 1 Session 1. The difference between their corresponding mutual minutia pairs, d4 and d3 is 8 pixels. The matching score of (c) with (a) using VeriFinger is 42. (b) and (d) show the corresponding unwrapped 3D fingerprints of impressions (a) and (c) respectively. The matching score of (d) with (b) is 63 and the difference between their corresponding mutual minutia pairs, d4 and d3 is 8 pixels.}
\label{fig::FlattenedVSUnwrapped}
\end{center}
\end{figure}

Figure~\ref{fig::FlattenedVSUnwrapped} (a) shows flattened impression 8-5 from Database 2 Session 1 and (c) shows flattened impression 8-2 from Database 1 Session 1. Both impressions were captured from the same person in different sessions. It can be observed that the two impressions were captured with different orientations. The matching score of (c) with (a) by using VeriFinger is 42. Each of these images also displays a pair of mutual minutiae, with their respective distances labeled as d1 in (a), and d3 in (c). The difference between the distance in the two images is 12 pixels. Ideally, we expect to have the same distance for mutual minutia pairs in two impressions. Figure~\ref{fig::FlattenedVSUnwrapped} (b) and (d) show the corresponding unwrapped 3D fingerprints of impressions (a) and (c) respectively. The matching score of (d) with (b) by using VeriFinger is 63 and the difference between their corresponding mutual minutia pairs, d4 and d3 is 8 pixels which is less than the distance difference in the flattened pair. A smaller difference in the distances between corresponding mutual minutiae pairs generally leads to a higher matching score, as it indicates better alignment between the fingerprint impressions.

\begin{figure}[!t]
\begin{center}
\begin{minipage}{0.24\textwidth}
\begin{center}
\includegraphics[width=1\textwidth]{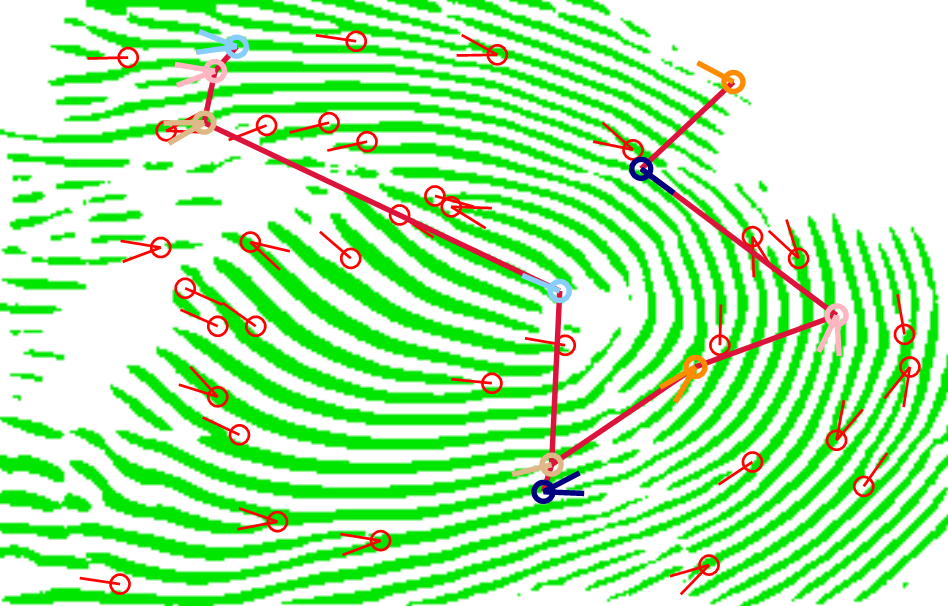}\\ 
(a) DB2-S1-8-5-Flattened
\end{center}
\end{minipage} \hfil
\begin{minipage}{0.24\textwidth}
\begin{center}
\includegraphics[width=1\textwidth]{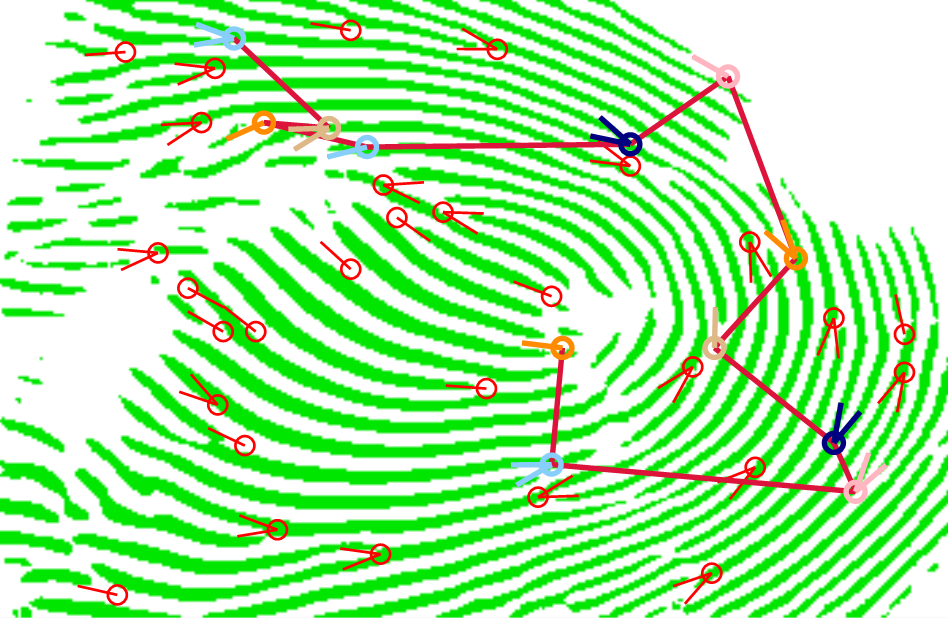}\\
(b) DB2-S1-8-5-Unwrapped
\end{center}
\end{minipage}
\begin{minipage}{0.24\textwidth}
\begin{center}
\includegraphics[width=1\textwidth]{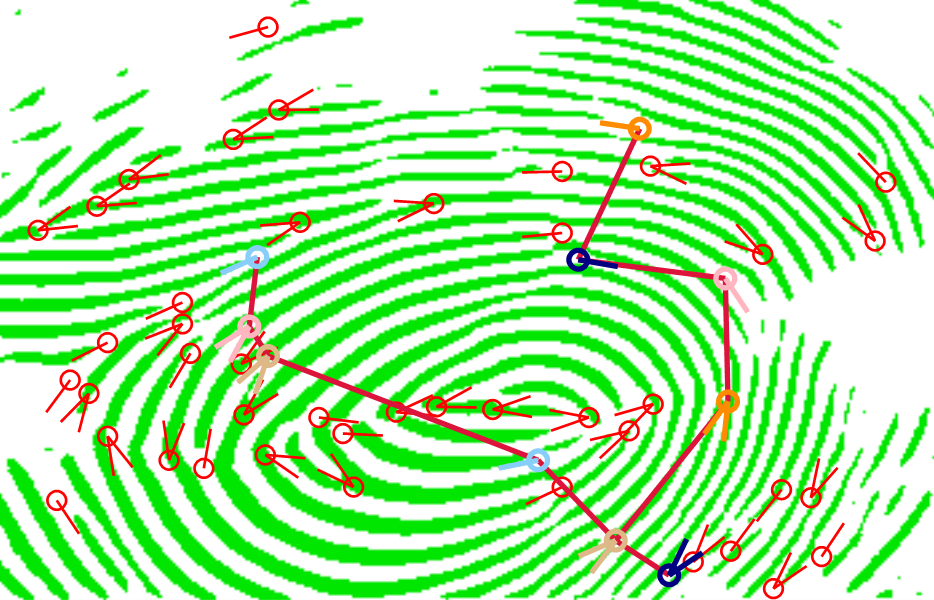}\\
(c) DB2-S2-8-2-Flattened
\vspace{2mm}
\par(a) vs (c)\par
matching score: 42
\end{center}
\end{minipage}
\begin{minipage}{0.24\textwidth}
\begin{center}
\includegraphics[width=1\textwidth]{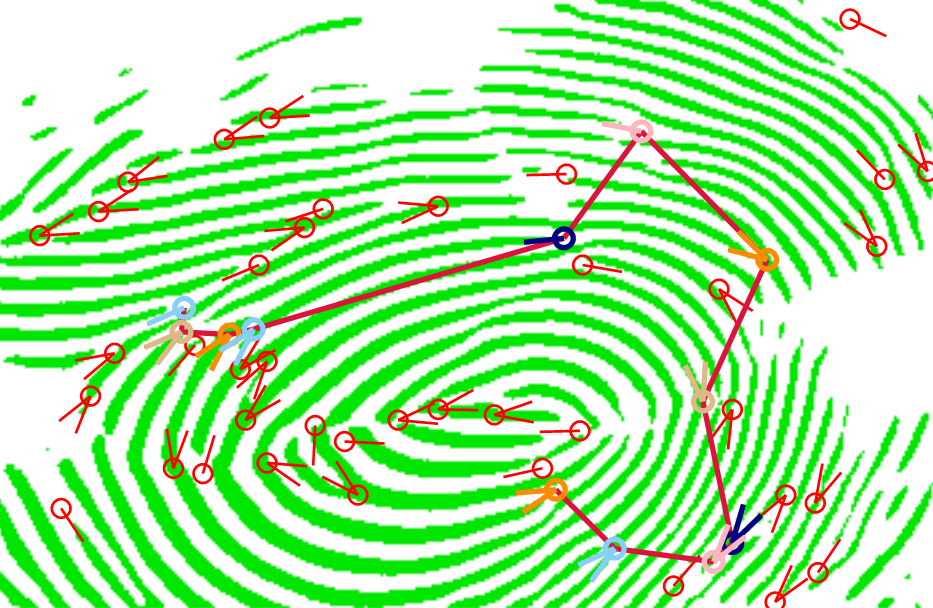}\\
(d) DB2-S2-8-2-Unwrapped
\vspace{2mm}
\par(b) vs (d)\par
matching score: 63
\end{center}
\end{minipage}
\caption{(a) and (c) show the minutiae detected by VeriFinger in the flattened fingerprints from Figure~\ref{fig::FlattenedVSUnwrapped} (a) and (c) respectively, with a total of 10 minutiae contributing to a matching score of 42. (b) and (d) display the minutiae detected in the unwrapped fingerprints from Figure~\ref{fig::FlattenedVSUnwrapped} (b) and (d) respectively. The total number of detected matched minutiae in this case is 12, which is higher than in the flattened case, contributing to a matching score of 63.}
\label{fig::FlattenedVSUnwrapped_Minutia}
\end{center}
\end{figure}

Figure~\ref{fig::FlattenedVSUnwrapped_Minutia} (a) and (c) show the minutiae detected by VeriFinger in the flattened fingerprints from Figure~\ref{fig::FlattenedVSUnwrapped} (a) and (c) respectively, with a total of 10 minutiae contributing to a matching score of 42. Figure~\ref{fig::FlattenedVSUnwrapped_Minutia} (b) and (d) display the minutiae detected in the unwrapped fingerprints from Figure~\ref{fig::FlattenedVSUnwrapped} (b) and (d) respectively. The total number of detected matched minutiae in this case is 12, which is higher than in the flattened case, contributing to a matching score of 63. These results indicate that the proposed unwrapping method enhances matching performance, especially when database impressions are captured at different angles and orientations.

It is also worth mentioning that point clouds in Database 2 in both sessions are the cropped sections from the center of a 3D fingerprint i.e. Regin of Interest (ROI). As a result, the global curvature in this database is moderate. We expect to have a more significant performance improvement by applying the unwrapping method on full-size 3D point clouds that include impressions with different registrations but at the moment there is no database with this characteristic available to verify this claim.

\begin{figure}[!t]
\begin{center}
\begin{minipage}{0.24\textwidth}
\begin{center}
\includegraphics[width=1\textwidth]{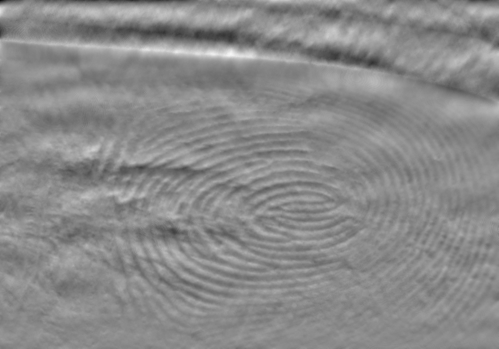}\\ 
\vspace{5mm}
(a) DB1-172-1-Flattened
\end{center}
\end{minipage} \hfil
\begin{minipage}{0.24\textwidth}
\begin{center}
\includegraphics[width=1\textwidth]{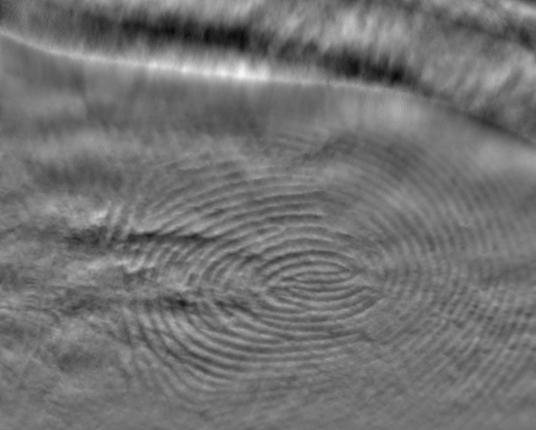}\\
(b) DB1-172-1-Unwrapped
\end{center}
\end{minipage}
\begin{minipage}{0.24\textwidth}
\begin{center}
\vspace{5mm}
\includegraphics[width=1\textwidth]{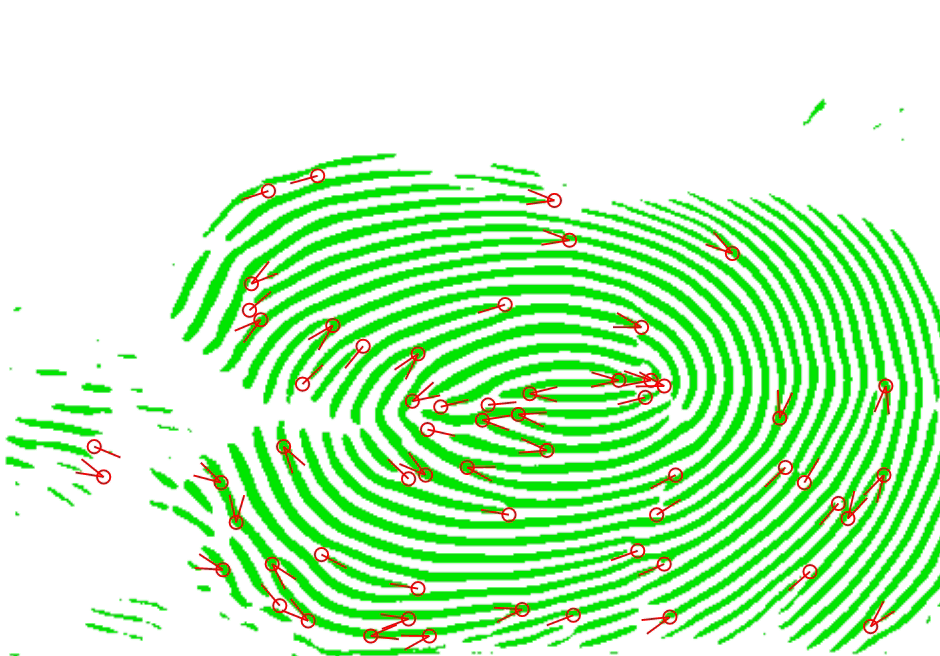}\\
(c) DB1-172-1-Flattened
\end{center}
\end{minipage}
\begin{minipage}{0.24\textwidth}
\begin{center}
\includegraphics[width=1\textwidth]{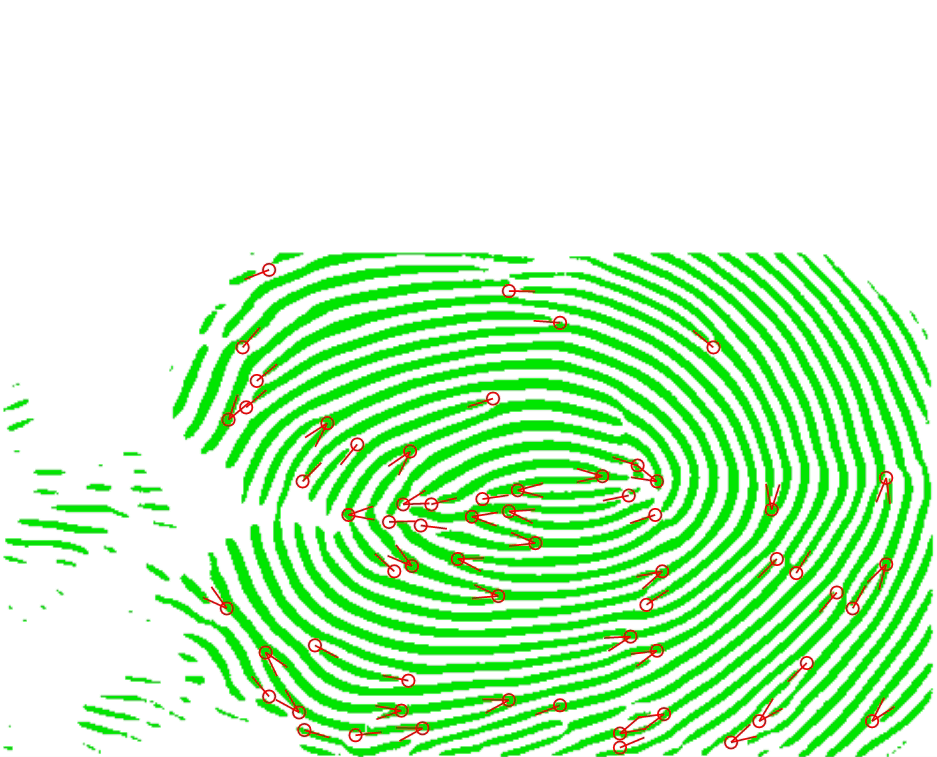}\\
(d) DB1-172-1-Unwrapped
\end{center}
\end{minipage}
\caption{(a) and (b) show respectively the generated flattened and unwrapped fingerprint images from a sample point cloud which includes data pints beyond the fingertip boundary. The detected minutia from these images are displayed in (c) and (d) respectively.}
\label{fig::FlattenedVSUnwrapped_distortion}
\end{center}
\end{figure}

The unwrapped fingerprint image generated from a full-size 3D point cloud may exhibit some distortion near the edges, particularly when the point cloud sample includes areas beyond the fingertip boundary. However, this distortion has minimal to no impact on the detected minutiae in the critical regions of the fingertip surface. Figure~\ref{fig::FlattenedVSUnwrapped_distortion} (a) and (b) shows respectively the generated flattened and unwrapped fingerprint images from a sample point cloud which includes data pints beyond the fingertip boundary. The detected minutia from these images are displayed in (c) and (d) respectively. 

\section{Conclusion and future work}\label{sec:Conclusion}
This paper proposed a method for generating grayscale images by unwrapping 3D point clouds, enabling 3D fingerprint recognition and identification solely from their respective 3D point clouds. The proposed method was evaluated in various scenarios. Its performance was assessed using two databases with varying point cloud sizes, one covering the entire fingertip surface and the other containing only the central region. These datasets include point clouds with different levels of acquisition accuracy. Additionally, experiments were conducted on datasets collected in different sessions to evaluate the method’s effectiveness in addressing registration limitations. The results indicated that the proposed method could resolve the registration issue in 3D fingerprint recognition and identification. The proposed method was also able to achieve a higher performance in 3D fingerprint recognition than the existing methods and competitive identification results with 3D point cloud flattening method and higher identification results that the rest of the available 3D fingerprint identification methods.

In future work, we will focus on further enhancing the method’s efficiency in handling registration challenges. While the current processing time for generating an unwrapped fingerprint does not hinder practical implementation, reducing the time cost would be preferable and will be another focus of our research.


\bibliographystyle{IEEEtran}
\bibliography{References}

\vfill\pagebreak

\end{document}